\documentclass[10pt,twocolumn,letterpaper]{article}

\usepackage{cvpr}
\usepackage{times}
\usepackage{epsfig}
\usepackage{graphicx}
\usepackage{amsmath}
\usepackage{amssymb}
\usepackage{algorithm}
\usepackage{multirow}
\usepackage{algorithmic}
\usepackage{mathrsfs}
\usepackage{mathtools}
\usepackage{subfigure}
\usepackage{enumitem}
\usepackage{tensor}
\usepackage{xcolor}
\usepackage{booktabs}
\usepackage{balance}
\usepackage{caption}
\usepackage[normalem]{ulem}
\algsetup{linenosize=\small}
\pdfoutput=1

\usepackage[pagebackref=true,breaklinks=true,letterpaper=true,colorlinks,bookmarks=false]{hyperref}
\renewcommand{\algorithmiccomment}[1]{\bgroup\hfill//~#1\egroup}

\cvprfinalcopy % *** Uncomment this line for the final submission

\def\cvprPaperID{1350} % *** Enter the CVPR Paper ID here

\ifcvprfinal\fi
\begin{document}

\setlength{\abovedisplayskip}{3pt}
\setlength{\belowdisplayskip}{3pt}
%%%%%%%%% TITLE
\title{Scalable Dense Non-rigid Structure-from-Motion: A Grassmannian Perspective}

\author{Suryansh Kumar$^{1}$\quad Anoop Cherian$^{1,2,4}$\quad Yuchao Dai$^{1, 3}$\quad Hongdong Li$^{1,2}$\\
${^1}$Australian National University, $^2$ACRV, 
$^3$NPU China, $^4$MERL Cambridge, MA
\\
{\tt\small firstname.lastname@anu.edu.au}\\
}

\maketitle
%\thispagestyle{empty}

%%%%%%%%% ABSTRACT
% \begin{abstract}
% Recently, clustering based approaches \cite{kumar2016multi} \cite{kumar2017spatio} to procure reconstruction of deformable object has demonstrated state-of-the-art results in non-rigid structure from motion (NRSfM) challenge \cite{challenge2017}. However, the proposed clustering based methods to NRSfM \cite{zhu2014complex}\cite{kumar2017spatio} is not scalable and therefore, dense reconstruction of the deforming shape cannot benefit from such an intrinsic idea. In addition, they depend on an unsubstantiated assumption that spatio-temporal deformation \emph{should} lie in the union of euclidean affine or linear subspace globally. In this paper, we took the idea of ``clustering inherently benefits reconstruction and vice-versa'' \cite{kumar2017spatio} to a noteworthy view-point by proposing a NRSfM algorithm which is scalable, which can handle noisy feature tracks, and exploits the inherent geometric structure of the deforming shapes on the Grassmann manifold. Extensive experiments on synthetic and real dense benchmark datasets shows the state-of-the-art performance.
% \end{abstract}
\begin{abstract} % A re-write of the abstract is needed to highlight it, the contribution and the novelty.
This paper addresses the task of dense non-rigid structure-from-motion (NRSfM) using multiple images. State-of-the-art methods to this problem are often hurdled by scalability, expensive computations, and noisy measurements. Further, recent methods to NRSfM usually either assume a small number of sparse feature points or ignore local non-linearities of shape deformations, and thus cannot reliably model complex non-rigid deformations. To address these issues, in this paper, we propose a new approach for dense NRSfM by modeling the problem on a Grassmann manifold. Specifically, we assume the complex non-rigid deformations lie on a union of local linear subspaces both spatially and temporally. This naturally allows for a compact representation of the complex non-rigid deformation over frames. We provide experimental results on several synthetic and real benchmark datasets. The procured results clearly demonstrate that our method, apart from being scalable and more accurate than state-of-the-art methods, is also more robust to noise and generalizes to highly non-linear deformations.

\end{abstract}

%%%%%%%%% BODY TEXT
\section{Introduction} \label{ss:intro}
% [From hongdong: Suryansh:  I have added some sentences in abstract;  I think you will need to cut short your Introduction section, making it short, and sharp, and strong.] \textcolor{blue}{Suryansh: Thanks a lot. Sure, I will address all the statements made by you. (In progress) }

%%%%%%%%%%%%%%%%%RECENT COMMENTS FROM HONGDONG%%%%%%%%%%%%%%%%%%%%%%%%%%%
% \textcolor{red}{Suryansh: please remove all footnotes, as that somehow suggest that you did not have time to incorporate those in the main body of your text, and your paper was finished in a rush.} \textcolor{blue}{Suryansh: Sure, I will keep this in mind}

% \textcolor{red}{Suryansh: I find equations (7)--(12) look dizzy. To remove/replace them with more concise/easier presentation , or refer it to Supplem. material ?   What do you think ? Some typos and (English) grammar errors in the comments in Algorithm-1. Also, please use the CVPR template predefined \ie \etc \etal , rather than creating your own style.   Do you have a graph or curve to explicitly show scalability of you method , \eg  computing time, memory consumption or footprint versus the number of points or the size of images ? \textcolor{blue}{Suryansh: Sure, I will refine my typos and grammar mistakes and will try to move Eq (7)-(12) in the supplementary material. Coming to the experiement part, yes I have done timing comparison with other methods but memory consumption experiment has not been performed.}} 
%%%%%%%%%%%%%%%%%%%%%%%%%%%%%%%%%%%%%%%%%%%%%
Non-rigid structure-from-motion (NRSfM) is a classical problem in computer vision, where the task is to recover the 3D shape of a deforming object from multiple images. Despite the fact that NRSfM for arbitrary deformation still remains an open problem, it can be solved efficiently under some mild prior assumptions about the deformation and the shape configuration \cite{dai2014simple,  lee2013procrustean, kumar2017spatio, zhu2014complex, akhter2009nonrigid,kumar2016multi, kumar2017monocular,gait,gait2}.

Even though the existing solutions to \emph{sparse NRSfM} have demonstrated outstanding results, they do not scale to dense feature points and their resilience to noise remains unsatisfactory.
Moreover, the state-of-the-art algorithms \cite{garg2013dense, dai2017dense} to solve \emph{dense NRSfM} are computationally expensive and rely on the assumption of global low-rank shape which, unfortunately, fails to cater to the inherent local structure of the deforming shape over time. Consequently, to represent dense non-rigid structure under such formulations seem rather flimsy and implausible.

%\footnote{Information about local structure is important in reliable representation, reconstruction and realization of minute deformations.}

\begin{figure}[t]
\begin{center}
\includegraphics[width=1.0\linewidth]{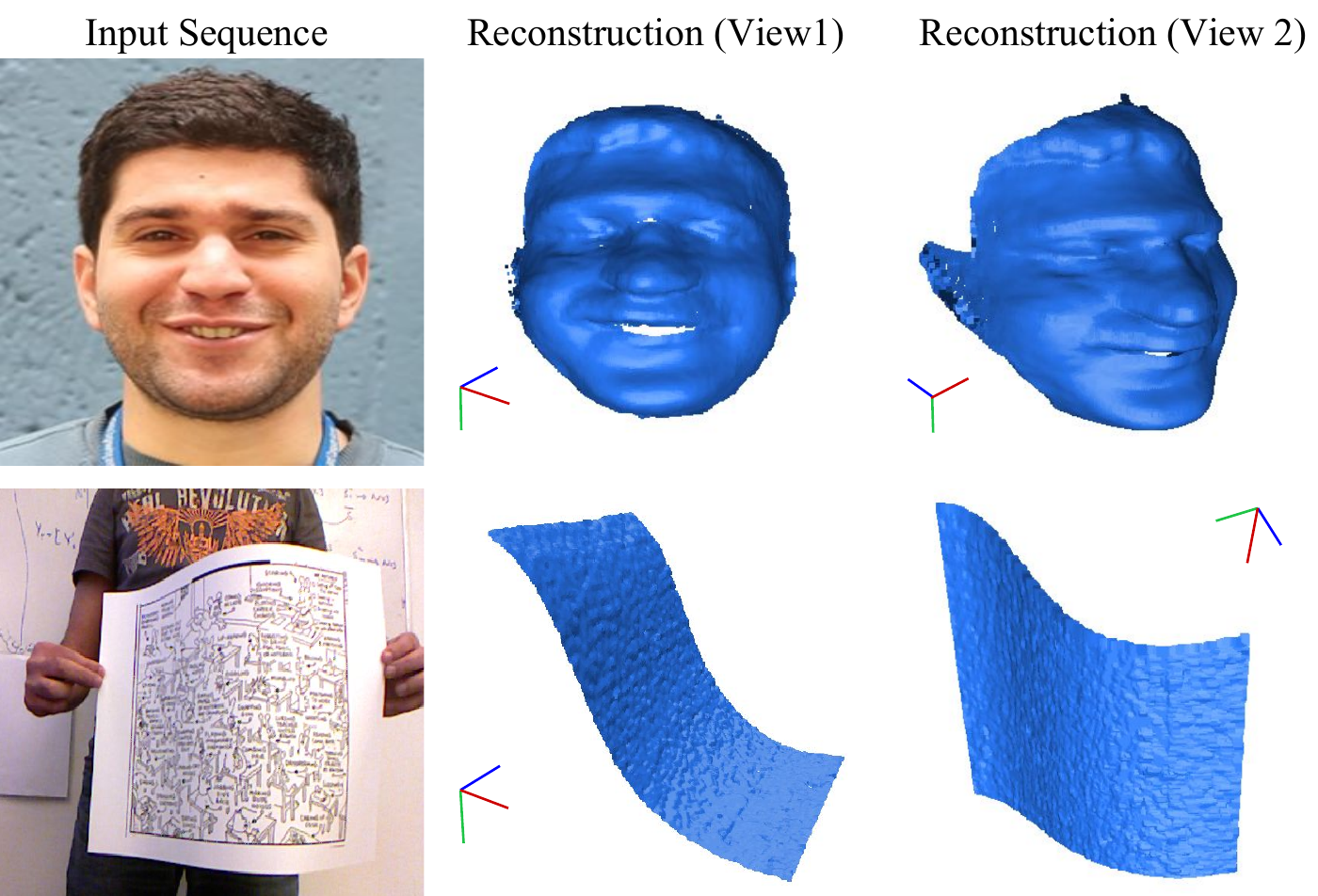}
\caption{\small{Our algorithm takes dense long-term 2D trajectories of a non-rigid deforming object as input, and provides a dense detailed 3D reconstruction of it. The reconstructed surface captures the complex non-linear motion which can be helpful for real world applications such as 3D virtual and augmented reality. Example frames are taken from publicly available real datasets: real face sequence\cite{garg2013dense} and kinect\_paper sequence\cite{varol2012constrained} respectively.} \label{fig:firstpageresult}}
\end{center}
\vspace{-0.8cm}
\end{figure}

For many real-world applications, such as facial expression reconstruction, limitations such as scalability, timing, robustness, reliable modeling, \etc, are of crucial concern. Despite these limitations, no template-free approach exists that can reliably deal with these concerns. In this paper, we propose a template-free dense NRSfM algorithm that overcomes these difficulties. As a first step to overcome these difficulties, we reduce the overall high-dimensional non-linear space spanned by representing the deforming shape as a union of several local low-dimensional linear subspaces. Our approach is based on a simple idea/assumption \ie, any complex deforming surface can be approximated by a locally linear subspace structure \cite{crane2013conformal}. We use this simple intuition in a spatio-temporal framework to solve dense NRSfM. This choice naturally leads to a few legitimate queries:

\noindent {\emph{ a) Why spatio-temporal framework for solving dense NRSfM?}} 
Spatio-temporal framework by Kumar \etal \cite{kumar2017spatio} has exhibited the state-of-the-art results on the recent NRSfM challenge \cite{challenge2017, jensen2018benchmark}. 
A recent method \cite{Agudo2017CVPR} which follows the same idea as proposed by Kumar \etal \cite{kumar2017spatio} has also observed an improvement in the reconstruction accuracy under such formulations.
% A recent method by Agudo \etal \cite{Agudo2017CVPR} has also advocated a similar observation under a different formulation. 
Even though the concept behind such a framework is elementary, no algorithm to our knowledge exists that exploit such an intrinsic idea for dense NRSfM. 

\noindent{\emph{ b) Why the previously proposed spatio-temporal methods are unable to handle dense NRSfM?}}
%{To formulate this intuition may seem easy, but it's not.  no need of this sentence. [hongdong]}
% Spatio-temporal modeling of NRSfM often requires a feedback from shape and trajectory space to local/global Euclidean subspace and vice-versa, which in fact is not straight-forward to formulate. However, in this work we achieve this by allowing the column permutation of trajectories in the classical NRSfM representation \cite{bregler2000recovering} and thorough clever use of algebraic concept such as Cholesky decomposition and Singular Value Decomposition to exploit the involved tensor representation \S \ref{ss:formulation}. This makes our formulation both effective and compact \footnote{Check supplementary material for details.}. 
% MAYBE REWRITE THIS PARAGRAPH IN a more authoriative way ==>   Our idea to use spatio-temporal framework to solve this problem may lead to few questions: \noindent {\emph{Why spatio-temporal framework for solving dense NRSfM?}} Recently, spatio-temporal framework has demonstrated the state-of-the-art results in NRSfM challenge \cite{challenge2017}. A few other recent methods \cite{zhu2014complex} \cite{Agudo2017CVPR} have also advocated a similar observation under a different formulation. Even though the concept behind such a framework is elementary, no algorithm exists to exploit such an intrinsic idea for dense NRSfM. \noindent{\emph{Why the previously proposed spatio-temporal formulation is unable to handle dense NRSfM?}}
The formulation proposed by Kumar \etal \cite{kumar2017spatio} and its adaptation \cite{Agudo2017CVPR} is inspired from SSC~\cite{elhamifar2009sparse}, and LRR~\cite{liu2013robust}. As a result, the complexity of their formulations grows exponentially in the order of the number of data points. This makes it difficult to solve dense NRSfM using their formulations. Moreover, these methods \cite{kumar2017spatio, zhu2014complex, Agudo2017CVPR} use an assumption that non-rigid shape should lie on a low-dimensional linear or affine subspace globally. In reality, such an assumption does not hold for all kinds of non-linear deformations \cite{wang2008manifold, rabaud2008re}. Although a recent spatio-temporal method proposed by Dai \etal \cite{dai2017dense} solves this task, it involves a series of least square problems to be solved, which is computationally demanding. %it very slow and complex. 

To overcome all these issues, we propose a spatio-temporal dense NRSfM algorithm which is free from such unwarranted assumptions and limitations. Instead, we adhere to the assumption that the low-dimensional linear subspace spanned by a deforming shape is locally valid. Such an assumption about shapes has been well studied in topological manifold theory \cite{absil2004riemannian,dollar2007non}. The Grassmann manifold is a topologically rich non-linear manifold, each point of which represents the set of all right-invariant subspaces of the Euclidean space. One property of the Grassmannian that is particularly useful in our setting is that the points in it can be embedded into the space of symmetric matrices. This property has been used in several computer vision applications that deal with subspace representation of data \cite{hamm2008grassmann,cetingul2009intrinsic}. Accordingly, in our problem, to model a non-linear shape, using a Grassmannian allows us to represent the shape as a set of ``smooth'' low-dimensional surfaces embedded in a higher dimensional Euclidean space. Such a representation not only reduces the complexity of our task but also makes our formulation robust and scalable as described below.

% \textcolor{blue}{Suryansh: I have to start refining from here} In contrast, our algorithm is free from such unwarranted assumptions and limitations. Instead we restrict the assumption of low-dimensional linear subspace to be valid locally which is well established for surface modeling \cite{dundas2002differential}. This allows us to represent the shape as a \AC{unclear} set of smooth low-dimensional manifolds (Grassmann manifold). \AC{Say something more about the Grassmann manifold here. How precisely it accomplishes this?} Such a representation not only reduces the \emph{computational?} complexity \S \ref{ss:Solution} of the problem, but also make our formulation robust and scalable \AC{as described below?} \S \ref{ss:Experiment and results}.
% \textcolor{blue}{Suryansh: Let me revisit the theory to come up with strong sentences and rewrite this part completely}
% \begin{figure}[t]
% \begin{center}
% \includegraphics[width=1.0\linewidth]{Figures/Intuition.pdf}
% \caption{\small{Example of topological manifold of dimension $n$ in general, which means that every point $p \in \mathcal{M}$ has a neighborhood that is homeomorphic to open subset of $\mathbb{R}^{n}$ (locally Euclidean).} \label{fig:intuition}}
% \end{center}
% \end{figure}

\noindent {\emph{c) Why Grassmann manifold?}}
It is well-known that the complex non-rigid deformations are composed of multiple subspaces that quite often fit a higher-order parametric model \cite{pasko2002function, sheng2011facial, zhu2014complex}. To handle such complex models globally can be very challenging -- both numerically and computationally. Consequently, for an appropriate representation of such a model, we decompose the overall non-linearity of the shape by a set of locally linear models that span a low-rank subspace of a vector space. As alluded to above, the space of all $d$-dimensional linear subspaces of $\mathbb{R}^{N}$ ($0 < d < N$) forms the Grassmann manifold \cite{absil2004riemannian, absil2009optimization}. Modeling the deformation on this manifold allows us to operate on the number of subspaces rather than on the number of vectorial data points (on the shape), which reduces the complexity of the problem significantly. Moreover, since each local surface is a low-rank subspace, it can be faithfully reconstructed using a few eigenvalues and corresponding eigenvectors, which makes such a representation scalable and robust to noise.

The aforementioned properties of the Grassmannian perfectly fit our strategy to model complex deformations, and therefore, we blend the concept of spatio-temporal representations with local low-rank linear models. This idea results in a two-stage coupled optimization problem \ie, local reconstruction and global grouping, which is solved efficiently using the standard ADMM algorithm \cite{boyd2011distributed}. As the local reconstructions are performed using a low-rank eigen decomposition, our representation is computationally efficient and robust to noise. We demonstrate the benefit of our approach to benchmark real and synthetic sequences \S\ref{ss:Experiment and results}. Our results show that our method outperforms previous state-of-the-art approaches by 1-2 \%  on all the benchmark datasets. Before we provide the details of our algorithm, we review some pertinent previous works in the next section.
\section{Background}
% [From hongdong: Please reduce this Section to no more than one column, or half page. this is important. ]
% [From yuchao: please consider using a table to complete these approaches such as used in Spatio-temporal ECCV 2014.]
% [From Yuchao: Another point, you may have too many footnotes.] \textcolor{blue}{Suryansh: In progress}
This section provides a brief background on the recent advancements in NRSfM, focusing mainly on the methods that are relevant to this work.

\par{\bf{Preliminaries:}} We borrow the notation system from Dai \etal's work \cite{dai2014simple} for its wide usage.  Given `$P$' feature points over `$F$' frames, we represent $W \in \mathbb{R}^{2F \times P}$, $S \in \mathbb{R}^{3F \times P}$,  $R \in \mathbb{R}^{2F \times 3F}$ as the measurement, the shape, and the rotation matrices, respectively. Here $R$ matrix is composed of block diagonal $R_{i} \in \mathbb{R}^{2 \times 3}$, representing per frame orthographic camera projection. Also, the notation $S^\sharp \in \mathbb{R}^{3P \times F}$ stands for the rearranged shape matrix, which is a linear mapping of $S$. We use $\|\ .\ \|_F$ and $\|\ .\ \|_*$ to denote the Frobenius norm and the nuclear norm, respectively.
%\AC{Is it SE(2) objects?} \Suryansh {yes non-rigid SFM are mostly solved under this setting}.
%Insert a table here show the 
{\footnotesize \begin{table*}[h!]
\centering
\begin{tabular}{|c|c|}
\hline
\begin{tabular}[c]{@{}c@{}}\label{tab:rw1} {(a) \bf{Dai \emph{et al.'s}}} \cite{dai2014simple}\\ $\underset{S^\sharp, E} {\text{minimize}} ~\| S^\sharp \|_* + \lambda \|E \|_F^2$ \\ ${\text{subject to:}}$ $W = RS + E$ \\ \end{tabular} & \begin{tabular}[c]{@{}c@{}} {(b) \bf{Zhu \emph{et al.'s} }}\cite{zhu2014complex}\\ $\underset{S^\sharp, C, E} {\text{minimize}} ~ \| C \|_* +  ~\gamma \| S^\sharp \|_* + \lambda \|E \|_1 $  \\ ${\text{subject to:}}$  $S^\sharp = S^\sharp C,  W = RS + E$ \\ \end{tabular}  \\ \hline
\begin{tabular}[c]{@{}c@{}} {(c) \bf{Kumar \emph{et al.'s}}} \cite{kumar2017spatio}\\ $\underset{S, S^\sharp, C_1, C_2} {\text{minimize}}\frac{1}{2} \|W-RS\|_F^2 + \lambda_1 \| C_1 \|_1 +  \lambda_2 \| S^\sharp \|_* + \lambda_3 \|C_2 \|_1$ \\ ${\text{subject to:}}$ $S = SC_1, S^\sharp = S^\sharp C_2, 1^{T}C_1 = 1^{T}, 1^{T}C_2 = 1^{T},$\\ $diag(C_1) = 0, diag(C_2) = 0,  S^\sharp = g(S)$ \\ \end{tabular} & \begin{tabular}[c]{@{}c@{}} {(d) \bf{Garg \emph{et al.'s}}}\cite{garg2013dense}\\ $\underset{S, R} {\text{minimize}}\frac{\lambda}{2} \|W-RS\|_F^2 + \sum_{f, i, p}\|\nabla S_{f}^{i}(p)\| + \tau \|S^\sharp\|_*$ \\ $\text{subject to:} $\\ $ R \in \mathbb{SO}(3)$\\ \end{tabular} \\ \hline
\end{tabular}
\caption{\small A brief summary of formulation used by some of the recent approaches to solve sparse and dense NRSfM which are closely related to our method. Among all these four methods only Garg \emph{et al.'s} \cite{garg2013dense} approach is formulated particularly for solving dense NRSfM. }
\label{tab:related works}
\vspace{-0.2cm}
\end{table*}}

\subsection{Relevant Previous Work}
\noindent \par{\it{Dai et al.'s approach:}} Dai \etal proposed a simple and elegant solution to NRSfM \cite{dai2014simple}. The work, dubbed ``prior-free", provides a practical solution as well as new theoretical insights to NRSfM. Their formulation involves nuclear norm minimization on $S^\sharp$ instead of $S$ --see Table \ref{tab:related works}(a). This is enforced due to the fact that $3K$ rank bound on $S$ is weaker than $K$ rank bound on $S^\sharp$, where $K$ refers to the rank of $S$. Although this elegant framework provides robust results for the shapes that span a single subspace, it may perform poorly on complex non-rigid motions \cite{zhu2014complex}. 
%Mathematically, minimizing over the convex envelope of $S^\sharp$ that is composed of multiple subspace is a loose constraint to reconstruct complex deformation \cite{zhu2014complex}. 

\noindent \par{{\it{Zhu et al.'s approach:}} To achieve better 3D reconstructions on complex non-rigid sequences, this work capitalized on the limitations of Dai \etal's work\cite{dai2014simple} by exploiting the union of subspaces in the shape space \cite{zhu2014complex}. The proposed formulation is inspired by LRR \cite{liu2013robust} in conjunction with Dai \etal work --see Table \ref{tab:related works}(b). In the formulation, $C \in \mathbb{R}^{F \times F}$, $E \in \mathbb{R}^{2F \times P}$ are the coefficient and error matrices. 

\noindent \par{\it{Kumar et al.'s approach:}} Kumar \etal exploits multiple subspaces both in the trajectory space and in the shape space \cite{kumar2017spatio}. This work demonstrated empirically that procuring multiple subspaces in the trajectory and shape spaces provide better reconstruction results. They proposed a joint segmentation and reconstruction framework, where segmentation inherently benefits reconstruction and vice-versa --see Table \ref{tab:related works}(c). In their formulation $C_1 \in \mathbb{R}^{P \times P}$, $C_2 \in \mathbb{R}^{F \times F}$ are the coefficient matrices and, $g(.)$ linearly maps $S$ to $S^\sharp$.

\noindent \par{\it{Dense NRSfM approach:}} Garg \etal developed a variational approach to solve dense NRSfM \cite{garg2013dense}. The optimization framework proposed by them employs total variational constraint on the deforming shape ($\nabla S_{f}^{i}(p)$) to allow edge preserving discontinuities, and trace norm constraints to penalize the number of independent shapes --see Table \ref{tab:related works}(d). Recently, Dai{\emph{ et al.}} has also proposed a dense NRSfM algorithm with a spatio-temporal formulation \cite{dai2017dense}.
%Consequently, it resulted in a biconvex formulation to the problem. The solution to this framework is computationally very expensive and probably needs GPU to speed up the implementation. but it's very slow due to computationally expensive gradient term in the formulation
\subsection{Motivation}
This work is intended to overcome the shortcomings of the previous approaches to solve dense NRSfM. Accordingly, we would like to outline the critical limitations associated with them. Although some of them are highlighted before, we reiterate it for the sake of completeness.

\begin{enumerate}[nolistsep, itemindent=0em, label=(\alph*)]
\item To solve dense NRSfM using the formulation proposed by Kumar \etal \cite{kumar2017spatio} and Zhu \etal \cite{zhu2014complex} is nearly impossible due to the complexity of their formulation \S \ref{ss:intro}.
Also, the error measure used by them is composed of Euclidean norm defined on the original data (see Table \ref{tab:related works}), which is not proper for non-linear data with a manifold structure \cite{absil2004riemannian, wang2015low}.

\item The algorithm proposed by Garg \etal \cite{garg2013dense} results in a biconvex formulation, which is computationally expensive and needs a GPU to speed up the implementation. Similarly, Dai \etal's recent work\cite{dai2017dense} is computationally expensive as well due to costly gradient term in their formulation.

\item Methods such as \cite{yu2015direct, li2009robust} rely on the template prior for dense 3D reconstruction of the object. Other piecewise approach for solving dense NRSfM \cite{russell2012dense} require a post-processing step to stitch all the local reconstructions.
\end{enumerate}

To avoid all the aforementioned limitations, we propose a new dense NRSfM algorithm. The primary contributions of this paper are as follows:
\begin{enumerate}[nolistsep]
\item A scalable spatio-temporal framework on the Grassmann manifold to solve dense NRSfM which does not need any template prior.
\item An effective framework that can handle non-linear deformations even with noisy trajectories and provides state-of-the-art results on benchmark datasets.
\item An efficient solution to the proposed optimization based on the ADMM procedure \cite{boyd2011distributed}.
\end{enumerate}

% can be solved efficiently on a regular CPU based desktop with more accurate reconstruction results. Although the clustering based approaches works quite well for general deformation with sparse set of feature points, it fail for dense non-linear deformation. As mentioned before, this limitation stems from the clustering framework used by the previous approaches \cite{kumar2017spatio} \cite{zhu2014complex}, and therefore holistic approach to rank minimization without any inference from local deformation may not be a suitable choice.

% Moreover, the clustering framework used by these approaches suffers from scalability issue.

% Few methods that uses local information such as rigidity to exploit NRSfM \cite{taylor2010non}, \cite{li2010multi} \cite{kumar2017monocular} has also been proposed.

\section{Problem Formulation}
In this section, we first provide a brief introduction to the Grassmann manifold and a suitable definition for a similarity distance metric on it, before revealing our formulation.

\subsection{Grassmann Manifold}
The Grassmann manifold, usually denoted as $\mathcal{G}(n, r)$, consists of all $r$-dimensional linear subspaces of $\mathbb{R}^{n}$, where $n>r$. A point on the Grassmann manifold is represented by a $n \times r$ matrix (say $X$), whose columns are composed of orthonormal basis of the subspace spanned by $X$, denoted as $\mathbf{span}(X)$ or in short as [$X$]. Let's suppose [$X_1$], [$X_2$] are two such points on this manifold, then among several similarity distances known for this manifold \cite{hamm2008grassmann}, we will be using the \emph{projection metric} distance given by $\small{d_g([X_1], [X_2]) = \frac{1}{\sqrt[]{2}} \|X_1X_1^{T}- X_2X_2^{T}\|_F}$, as it allows directly embedding the Grassmannian points into a Euclidean space (and the use of the Frobenius norm) using the mapping $X\rightarrow XX^T$. 
% \begin{equation}\label{eq:GMM}
% \begin{small}
% d_g([X_1], [X_2]) = \frac{1}{\sqrt[]{2}} \|X_1X_1^{T}- X_2X_2^{T}\|_F
% \end{small}
% \end{equation}
%While a number of different metric have been proposed for the Grassmann manifold, $d_g$ provides the projection distance. 
With this metric, $(\mathcal{G}, d_g)$ forms a metric space. Interested readers may refer to \cite{hamm2008grassmann} for details.

\begin{figure*}
\centering
\includegraphics[width=0.95\textwidth, height=0.25\textheight] {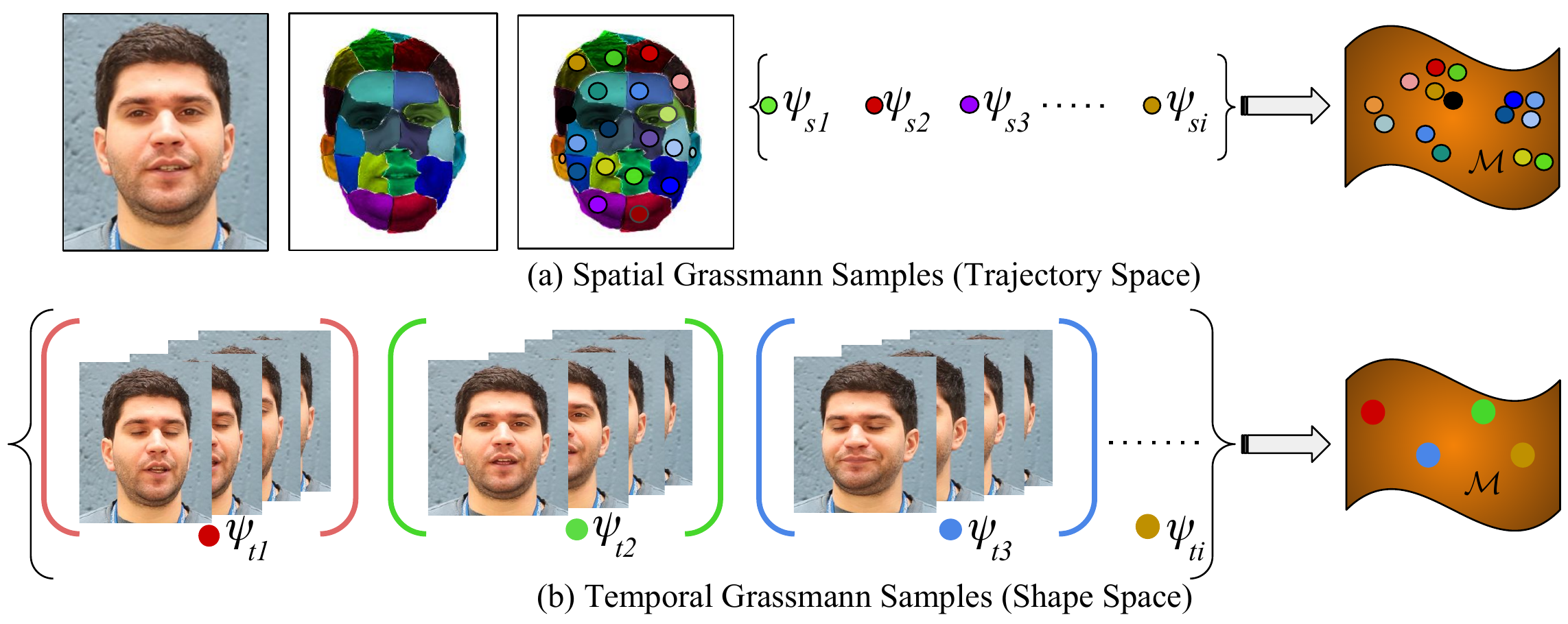}~~~
\caption{\small{Conceptual illustration of data point representation on the grassmann manifold. Each local subspace can equivalently be represented by a single point on the manifold. {\bf{Top row:}} Construction of grassmann samples in the trajectory space using spatial information. {\bf{Bottom row:}}  Construction of grassmann samples in the shape space by partitioning the shapes in a sequential order over frames.}}
\label{fig:manifoldconcept}
\end{figure*}

\subsection{Formulation}\label{ss:formulation}
% In this section, we share the details of our algorithm to solve dense NRSfM under orthographic projection. We start our discussion with the classical representation to NRSfM \ie
With the relevant background as reviewed in the above sections, we are now ready to present our algorithm to solve the dense NRSfM task under orthographic projection. We start our discussion with the classical representation to NRSfM \ie,
\begin{equation}
\begin{small}
W_{s} = RS_{s}\label{eq:reprojectionErr}
\end{small}
\end{equation}
where, $W_{s} \in \mathbb{R}^{2F \times P}$, $R = {\text{blkdiag}}(R_1, ..., R_F) \in \mathbb{R}^{2F \times 3F}$, $S_{s} \in \mathbb{R}^{3F \times P}$. The motive here is, given the input measurement matrix, solve for rotation ($R$) and 3D shape ($S_s$). To serve this objective, Eq.\eqref{eq:reprojectionErr} maintains the camera motion and the shape deformation such that it complies with the image measurements. For our method, we solve for rotations using the Intersection method \cite{dai2014simple} by assuming that the multiple non-rigid motions within a single deforming object, over frames, can be faithfully approximated by per frame single relative camera motion with a higher rank\footnote{Check the supplementary material for a detail discussion on rotation.}. Accordingly, our goal reduces to develop a systematic approach that can reliably explain the non-rigid shape deformations and provides better 3D reconstruction. We use  subscript `$s$' in Eq.\eqref{eq:reprojectionErr} to indicate that the column permutations of $S_s$ and $W_s$ matrix are allowed. However, the  column permutations of $S_t^{\sharp}$ is inadmissible as it results in discontinuous trajectories over frames. 

\noindent{\bf{Grassmannian Representations in Trajectory Space:}}
Let's suppose $\mathbf{\Psi}_s$ = $\{ {\it{\Psi_{s1}}}, {\it{\Psi_{s2}}}, .. {\it{\Psi_{sK_s}}}\}$ is the set of points on the Grassmann manifold generated using $S_s$ matrix, then ${\bf{\mathcal{T}}}_s = \big\{ (\mathit{\Psi_{s1}})(\mathit{\Psi_{s1}})^T, (\mathit{\Psi_{s2}})(\mathit{\Psi_{s2}})^T..., (\mathit{\Psi_{sK_s}})(\mathit{\Psi_{sK_s}})^T \big\}$ represents a tensor which is constructed by mapping  all symmetric matrices of the Grassmann data points ---refer Figure \ref{fig:manifoldconcept}(a). As discussed before in \S \ref{ss:intro}, to explain the complex deformations, we reduce the overall non-linear space as a union of several local low-dimensional linear spaces which form the sample points on the Grassmann manifold. But, the notion of self-expressiveness is valid only for Euclidean linear or affine subspace. To apply self-expressiveness on the Grassmann manifold one has to adopt linearity onto the manifold. Since, Grassmann manifold is isometrically equivalent to the symmetric idempotent matrices \cite{chikuse2012statistics}, we embed the Grassmann manifold into the symmetric matrix manifold, where the self-expression can be defined in the embedding space. This leads to the following optimization:
% \par{\bf{Grassmannian representation in trajectory space:}}
% The notion of self-expressiveness is valid for euclidean linear or affine subspace. Consequently, its inadequate to  use the
% euclidean metric to measure the trajectory reconstruction error on a surface with underlying manifold structure. To apply self-expressiveness on the Grassmann manifold one has to adopt linearity onto the manifold. Since, Grassmann manifold is isometrically equivalent to the symmetric idempotent matrices subspace \cite{chikuse2012statistics}. Hence, to use self-expression in the abstract Grassmann, we embed the Grassmann manifold into the symmetric matrix manifold, where the self-expression can be defined in the embedding space. Let's suppose $\mathbf{\Psi_{s}}$ = $\{ {\it{\Psi_{s1}}}, {\it{\Psi_{s2}}}, .. {\it{\Psi_{sK_s}}}\}$ is the set of points on the Grassmann manifold generated using $S_s$ matrix, then ${\bf{\mathcal{T}_s}} = \big\{ (\mathit{\Psi_{s1}})(\mathit{\Psi_{s1}})^T, (\mathit{\Psi_{s2}})(\mathit{\Psi_{s2}})^T..., (\mathit{\Psi_{sK_s}})(\mathit{\Psi_{sK_s}})^T \big\}$ represents a tensor which is constructed by mapping  all symmetric matrices of the Grassmann data points. $C_s \in \mathbb{R}^{K_s \times K_s}$ denotes the coefficient matrix in the trajectory space with $K_s$ representing the total number of spatial clusters ---refer Figure \ref{fig:manifoldconcept}. This leads to the following optimization:
\begin{equation}
\begin{small}
\begin{aligned}
& \displaystyle \underset{E_s, C_s}  {\text{minimize}} ~\|E_s\|_{F}^2 + \lambda_1\|C_s\|_* \\
& \displaystyle \text{\it{subject to:}} ~{\bf{\mathcal{T}}}_s = {\bf{\mathcal{T}}}_s C_s + E_s
\end{aligned}
\end{small}
\end{equation}
We denote $C_s \in \mathbb{R}^{K_s \times K_s}$ as the coefficient matrix with `$K_s$' as the total number of spatial groups. Here, $E_s$ measures the trajectory group reconstruction error as per the manifold geometry. Also, we would like to emphasize that since the object undergoes deformations in the 3D space, we operate in 3D space rather than in the projected 2D space. $\|~\|_*$ is enforced on $C_s$  for a low-rank solution.

\noindent{\bf{Grassmannian Representations in Shape Space:}}
%\AC{Non-rigid objects perform distinct temporal activities over time.} Suryansh: Thank you so much for this comment.
Deforming object attains different state over time which adheres to distinct temporal local subspaces \cite{kumar2017spatio}. Assuming that the temporal deformation is smooth over-time, we express deforming shapes in terms of local self-expressiveness across frames as:
\begin{equation}
\begin{small}
\begin{aligned}
& \displaystyle \underset{E_t, C_t}  {\text{minimize}} ~\|E_t\|_{F}^2 + \lambda_2\|C_t\|_* \\
& \displaystyle \text{\it{subject to:}}
{\bf{\mathcal{T}}}_t = {\bf{\mathcal{T}}}_t C_t + E_t
\end{aligned}
\end{small}
\end{equation}
Similarly, ${\bf{\mathcal{T}}}_t$ is the set of all symmetric matrices constructed using a set of Grassmannian samples $\mathbf{\Psi}_t $, where $\mathbf{\Psi}_t$ contains the samples which are drawn from $S_{t}^{\sharp} \in \mathbb{R}^{3P \times F}$  ---refer Figure \ref{fig:manifoldconcept}(b). Intuitively, $S_{t}^{\sharp}$ is a shape matrix with each column as a deforming shape. $E_t$, $C_t \in \mathbb{R}^{K_t \times K_t}$ represent the temporal group reconstruction error and coefficient matrix respectively, with $K_t$ as the number of temporal groups. $\|~\|_*$ is enforced on $C_t$  for a low-rank solution.

\noindent{\bf{Spatio-Temporal Formulation:}} Combining the above two objectives and their constraints with reprojection error term give us our formulation. Our representation blends the local subspaces structure along with the global composition of a non-rigid shape. Thus, the overall objective is:
%\AC{the eqn number (4) is misaligned.} I will align it
\begin{equation} \label{eq:overallOptimization} 
\begin{small}
\begin{aligned}
& \displaystyle \underset{S_{s}, S_{t}^{\sharp}, E_s, E_t, C_s, C_t} {\text{minimize}} \hspace{-0.3cm} \mathbf{E} = \frac{1}{2} ~\|W_{s}-RS_{s}\|_F^2 + \gamma \| S_{t}^{\sharp} \|_* +  \lambda_1\|E_s \|_F^2 \\
& \displaystyle + \lambda_2\|E_{t}\|_F^{2} + \lambda_3 \|C_s\|_* +  \lambda_4 \|C_t\|_*\\
& \displaystyle \text{\it{subject to:}} ~{\bf{\mathcal{T}}}_s = {\bf{\mathcal{T}}}_s C_s + E_s; {\bf{\mathcal{T}}}_t = {\bf{\mathcal{T}}}_t C_t + E_t;\\
& \displaystyle \hspace{10ex}\mathbf{\Psi}_s  = \mathit{\xi(C_s, S_s, \sigma_{\epsilon})}; \mathbf{\Psi}_t  = \mathit{\xi(C_t, S_t^{\sharp}, \sigma_{\epsilon})}; \\
& \displaystyle \hspace{10ex} S_{s} = \zeta\big(\mathbf{\Psi}_s, \Sigma_s, V_s, N_s); S_{t}^{\sharp} = \zeta\big( \mathbf{\Psi}_t, \Sigma_{t}^{\sharp}, V_t, N_t ); \\
& \displaystyle \hspace{10ex} S_{t}^{\sharp} = \mathscr{T}_1(S_{s});  W_s = \mathscr{T}_2(W_{s}, S_s);
\end{aligned}
\end{small}
\end{equation}
The re-projection error constraint performs the 3D reconstruction using  $W_{s}$ and $R$. Meanwhile, the local subspace grouping naturally enforces the union of subspace structure in $S_s$, $ S_{t}^{\sharp}$ with corresponding low-rank representations of the coefficient matrices $C_s$ and $C_t$. Here, the function $\xi(.)$ draws inference from $C$ matrices to refine Grassmannian sample set, both in trajectory and shape spaces. The function $\zeta(.)$ reconstructs $S_s$ and $S_{t}^{\sharp}$ matrices based on a set of local subspaces ($\mathbf{\Psi}_s, \mathbf{\Psi}_t, V_s, V_t $), singular values ($\Sigma_s$, $\Sigma_t$) and the number of top eigenvalues ($N_s, N_t$). The function $\mathscr{T}_1(.)$ transforms $S_{s} \in \mathbb{R}^{3F \times P}$ matrix to $S_{t}^{\sharp} \in \mathbb{R}^{3P \times F}$ matrix and $\mathscr{T}_2(.)$ function rearranges $W_s$ matrix as per the recent ordering of $S_s$\footnote{It's important to keep track of column permutation of $W_s$, $S_s$.}. Parameters such as `$\sigma_{\epsilon}$', `$N_s$' and `$N_t$' provides the flexibility to handle noise and adjust computations. Note that the element of the sets $\mathbf{\Psi}_s, \mathbf{\Psi}_t, V_s ~\text{and} ~V_t$ are obtained using SVD. The above equation \ie Eq: \eqref{eq:overallOptimization} is a coupled optimization problem where the solution to $S$ matrices influence the solution of $C$ matrices and vice-versa, and ${\small\mathscr{T}_1}()$ connects $ S_{t}^{\sharp}$ to $S_s$.
%\AC{say something on the coupling?}\textcolor{blue}{Suryansh: In progress, out for lunch}
%===================================
\begin{algorithm*}[t!]
\caption{Scalable Dense Non-Rigid Structure from Motion: A Grassmannian Perspective}\label{alg:Algorithm1}
\begin{algorithmic}[1]
\REQUIRE
$W_s$, $R$ using \cite{dai2014simple}, tuning parameters: $\lambda_1$, $\lambda_2$, $\lambda_3$, $\lambda_4$, $\gamma$, $\rho=1.1$, $\beta=1e^{-3}$, $\beta_m=1e^{6}$, $\epsilon = 1e^{-12}$, $K_s$, $K_t$.\\
\hspace{-0.3cm}{\bf Initialize:} $S_s$ = pseudoinverse($R$)$W_s$ and ${ S^{\sharp}_t} = \mathscr{T}_1(S_s)$.\\

\hspace{-0.3cm}{\bf Initialize:} `$K_t$' temporal data points on the Grassmann manifold using `${ S^{\sharp}_t}$' matrix,
$\mathbf{\Psi}_t$ = $\{\it{\Psi_{ti}}\}_{i=1}^{K_t}$.\\
\hspace{-0.3cm}{\bf Initialize:} `$K_s$' spatial data points on the Grassmann manifold using `${ S_s}$' matrix, $\mathbf{\Psi}_s$ = $\{\it{\Psi_{si}}\}_{i=1}^{K_s}$.\\
\hspace{-0.3cm}{\bf Initialize:} The auxiliary variables $J_s$, $J_t$ and Lagrange multiplier $\{Y_i\}_{i=1}^{3}$ as zero matrices.\\
\hspace{-0.3cm}{\bf Initialize:} $\Omega_{ij}^{s}$ = $trace[ \big({\it{\Psi_{sj}^T} } {\it{\Psi_{si}}} \big) \big({\it{\Psi_{si}^T} } {\it{\Psi_{sj}}} \big)]$,
%second line
$\Omega_{ij}^{t}$ = $trace[ \big({\it{\Psi_{tj}^T} } {\it{\Psi_{ti}}} \big) \big({\it{\Psi_{ti}^T} } {\it{\Psi_{tj}}} \big)]$, $\mathbf{\Omega}_s = (\Omega_{ij}^{s})_{i, j = 1}^{K_s}$, $\mathbf{\Omega}_t = (\Omega_{ij}^{t})_{i, j = 1}^{K_t}$\\

\hspace{1.2cm} $L_sL_s^{T}$ = Cholesky($\mathbf{\Omega}_s$), $L_tL_t^{T}$ = Cholesky($\mathbf{\Omega}_t$)
\vspace{0.2cm}\\
%\COMMENT ALM based solution to Equation \ref{eq:overallOptimization3}
\WHILE {not converged}
%\STATE \COMMENT SPATIAL VARIABLES
\STATE $S_s$ $\gets$ $\Big(\beta\big(\mathscr{T}_1^{-1}(S_t^{\sharp}) + \frac{\mathscr{T}_1^{-1}(Y_1)}{\beta}\big) + R^T W_s \Big)\big/ (R^{T}R + \beta I)$  \\

\STATE $C_s$ $\gets$ $\Big(2\lambda_1 L_s L_s^{T} + \beta(J_s -\frac{Y_2}{\beta}) \Big) $ $\big(2 \lambda_1 L_s L_s^{T} + \beta I_s  \big)^{-1}$\\

\STATE $\mathbf{\Psi}_s$ $\gets$ $\xi(C_s, S_s, \sigma_{\epsilon})$ \hspace{3.0cm} \COMMENT {Update spatial Grassmann points}

\STATE $S_s$ $\gets$ $\zeta(\mathbf{\Psi}_s, \Sigma_s, V_s, N_s)$; \hspace{2.0cm} \COMMENT {refine based on top $N_s$ eigen value}
\STATE $J_s$ $\gets$  $U_{J_s} \mathcal{S}_{\frac{\lambda_3}{\beta}}(\Sigma_{J_s}) V_{J_s}$, where $[U_{J_s}, \Sigma_{J_s}, V_{J_s}] = \emph{svd}(C_s + \frac{Y_2}{\beta})$ and $\mathcal{S}_\tau[x]$ = sign(x)max($|x|$-$\tau$, 0)
%\COMMENT TEMPORAL VARIABLES
\STATE $S_t^{\sharp}$ $\gets$ $U_t \mathcal{S}_{\frac{\gamma}{\beta}} (\Sigma_t)V_t$, where $[U_t, \Sigma_t, V_t] = \emph{svd}(\mathscr{T}_1(S_s) - \frac{Y_1}{\beta})$ and $\mathcal{S}_\tau[x]$ = sign(x)max($|x|$-$\tau$, 0)\\

\STATE $C_t$ $\gets$  $\Big(2\lambda_2 L_t L_t^{T} + \beta(J_t -\frac{Y_3}{\beta}) \Big) $ $\big(2 \lambda_2 L_t L_t^{T} + \beta I_t  \big)^{-1}$

\STATE $\mathbf{\Psi}_t$ $\gets$ $\xi(C_t, S_t^{\sharp}, \sigma_{\epsilon})$ \COMMENT {Update temporal Grassmann points}

\STATE $S_t^{\sharp}$ $\gets$ $\zeta(\mathbf{\Psi}_t, \Sigma_t, V_t, N_t)$;\hspace{2.0cm} \COMMENT {refine based on top $N_t$ eigen value} \\

\STATE $J_t$ $\gets$  $U_{J_t} \mathcal{S}_{\frac{\lambda_4}{\beta}}(\Sigma_{J_t}) V_{J_t}$, where $[U_{J_t}, \Sigma_{J_t}, V_{J_t}] = \emph{svd}(C_t + \frac{Y_3}{\beta})$ and $\mathcal{S}_\tau[x]$ = sign(x)max($|x|$-$\tau$, 0)

\STATE $\Omega_{ij}^{s}$ $\gets$ $trace[ \big({\it{\Psi_{sj}^T} } {\it{\Psi_{si}}} \big) \big({\it{\Psi_{si}^T} } {\it{\Psi_{sj}}} \big)]$,
%second line
$\Omega_{ij}^{t}$ $\gets$ $trace[ \big({\it{\Psi_{tj}^T} } {\it{\Psi_{ti}}} \big) \big({\it{\Psi_{ti}^T} } {\it{\Psi_{tj}}} \big)]$;\\

\STATE $\mathbf{\Omega}_s \gets (\Omega_{ij}^{s})_{i, j = 1}^{K_s}$, $\mathbf{\Omega}_t \gets (\Omega_{ij}^{t})_{i, j = 1}^{K_t}$;\COMMENT. $\mathbf{\Omega}_s \succeq 0,\mathbf{\Omega}_t \succeq 0$, if $\mathbf{\Omega}_s || \mathbf{\Omega}_t = 0$ add $\delta \mathbf{I}$ to make it  $\succ 0$ (see suppl. material)\\

\STATE $L_sL_s^{T}$ = Cholesky($\mathbf{\Omega}_s$), $L_tL_t^{T}$ = Cholesky($\mathbf{\Omega}_t$); \\

\STATE $W_s$ $\gets$ $\mathscr{T}_2(W_{s}, S_s)$ \hspace{2.0cm}\COMMENT {Note: Column permutation for $W_s$ and $S_s$ should be same.}

\STATE $Y_1$ := $Y_1 + \beta(S_{t}^{\sharp} - \mathscr{T}_1(S_{s}))$, $Y_2$ := $Y_2 + \beta(C_s - J_s)$, $Y_3$ := $Y_3 + \beta(C_t - J_t)$; \COMMENT Update Lagrange multipliers\\
\STATE $\beta\gets \text{minimum}(\rho\beta, \beta_m)$ \\
\STATE maxgap := maximum($[\| S_t^\sharp - \mathscr{T}_1(S_{s})\|_{\infty}, \|C_s - J_s \|_{\infty}, \|C_t - J_t\|_{\infty} ]$)
\IF{(maxgap $<$ $\epsilon$ $ \parallel \beta > \beta_m$)}
	\STATE break;
\ENDIF \COMMENT check for the convergence
\ENDWHILE \COMMENT Note: $\delta$ is a very small positive number and $\mathbf{I}$ symbolizes identity matrix.
\ENSURE $S_s$, $S_t$, $C_s$, $C_t$.
\COMMENT {Note: Kindly use economical version of {\it{svd}} on a regular desktop.}
\end{algorithmic}
\end{algorithm*}

%==============================

\section{Solution} \label{ss:Solution}
The formulation in Eq.\eqref{eq:overallOptimization} is a non-convex problem due to the bilinear optimization variables (${\bf{\mathcal{T}}}_s C_s$, ${\bf{\mathcal{T}}}_t C_t$), hence a global optimal solution is hard to achieve. However, it can be efficiently solved using Augmented Lagrangian Methods (ALMs) \cite{boyd2011distributed}, which has proven its effectiveness for many non-convex problems. Introducing Lagrange multipliers $(\{Y_i\}_{i=1}^3)$ and auxiliary variables $(J_s, J_t)$ to Eq.\eqref{eq:overallOptimization} gives us the complete cost function as follows:
\begin{equation}
\begin{small}
\begin{aligned}
& \displaystyle  \underset{S_{s}, S_{t}^{\sharp}, C_s, C_t, J_s, J_t} {\text{minimize}} \hspace{-0.3cm} \mathbf{E} = \frac{1}{2} ~\|W_{s}-RS_{s}\|_F^2 + \frac{\beta}{2}\| S_{t}^{\sharp} - \mathscr{T}_1(S_{s})\|_F^2 + \\
& \displaystyle <Y_1, S_{t}^{\sharp} - \mathscr{T}_1(S_{s})> + \gamma \| S_{t}^{\sharp} \|_* + \lambda_1\|{\bf{\mathcal{T}}}_s-{\bf{\mathcal{T}}}_s C_s\|_F^2 + \lambda_3 \|J_s\|_* \\
& \displaystyle + \frac{\beta}{2}\|C_s - J_s\|_F^2 + <Y_2, C_s - J_s> + \lambda_2\|{\bf{\mathcal{T}}}_t-{\bf{\mathcal{T}}}_t C_t\|_F^{2} + \\
& \displaystyle \lambda_4 \|J_t\|_* + \frac{\beta}{2}\|C_t -J_t \|_F^2 + <Y_3, C_t-J_t> \\
 & \displaystyle \text{\it{subject to:}} ~\mathbf{\Psi}_s  = \mathit{\xi(C_s, S_s, \sigma_{\epsilon})}; \mathbf{\Psi}_t  = \mathit{\xi(C_t, S_t^{\sharp}, \sigma_{\epsilon})}; \\
%& \displaystyle \hat{gs} = \xi(\mathbf{\Gamma_s}); \hat{gt} = \xi(\mathbf{\Gamma_t});   \\
& \displaystyle \hspace{10ex} S_{s} = \zeta(\mathbf{\Psi}_s,\Sigma_s, V_s, N_s);  S_{t}^{\sharp} = \zeta(\mathbf{\Psi}_t,  \Sigma_t^{\sharp}, V_t, N_t); \\
& \displaystyle \hspace{10ex} W_s = \mathscr{T}_2(W_{s}, S_s);
\end{aligned} \label{eq:overallOptimization3}
\end{small}
\end{equation}
The function $\xi(.)$ first computes the SVD of $C$ matrices, \ie $C = [U_c, \Sigma_c, V_c]$, then forms a matrix $A$ such that $A_{ij} = [XX^T]_{ij}^{\sigma_{\epsilon}}$, where $\sigma_{\epsilon}$ is set empirically based on noise levels and $X = U_c(\Sigma_c)^{0.5}$ (normalized). Secondly, it uses $A_{ij}$ to form new Grassmann samples from the $S$ matrices\cite{ji2014efficient}. Notice that $\xi(.)$ operates on $C$ matrices whose dimensions depend on the number of Grassmann samples.  This reduces the complexity of the task from exponential in the number of vectorial points to exponential in the number of linear subspaces. The later being of the order 10-50, where as the former can go more than 50,000 for dense NRSfM.

The $\zeta(.)$ function is defined as follows $\zeta$ = $\big\{ (\mathbf{\Psi}_a, \Sigma_a, V_a, r) | S_a$ = $\mathbf{horzcat}(\mathbf{\Psi}_a^r\Sigma_a^rV_a^r), \forall 1 \leq a \leq \mathbf{Card}(\mathbf{\Psi}_a), r \in \mathbb{Z}^{+} \big\}$, where $r$ stands for top-$r$ eigenvalues, $\mathbf{Card}(.)$ denotes the cardinal number of the set and $\mathbf{horzcat}(.)$ denotes for the horizontal concatenation of matrices. Intuitively, $\zeta(.)$ reconstructs back each local low-rank subspace. During implementation, we use $S_s$, $S_{t}^\sharp$ in place of $S_a$ accordingly. The optimization variables over iteration are obtained by solving for one variable at a time treating others as constant, keeping the  constraints intact. For detailed derivations for each sub-problem and proofs, kindly refer to the supplementary material. The pseudo code of our implementation is provided in {\bf{Algorithm \ref{alg:Algorithm1}}}.
%\AC{provide some insights to this definition of $\zeta$.} Done
%\AC{Any note on the computational complexity of the scheme?} \Suryansh : NOne for now. I will try to put it 

\section{Experiments and Results} \label{ss:Experiment and results}
We compare the performance of our method against four previously reported state-of-the-art approaches, namely Dense Spatio-Temporal DS \cite{dai2017dense}, Dense Variational DV \cite{garg2013dense}, Trajectory Basis PTA \cite{akhter2009nonrigid} and Metric Projection MP \cite{paladini2012optimal}. To test the performance, we used dense NRSfM dataset introduced by Garg\emph{ et al.} \cite{garg2013dense} and Varol\emph{ et al.} \cite{varol2012constrained} under noisy and noise free conditions. For quantitative evaluation of 3D reconstruction, we align the estimated shape $S_{est}^{t}$ with ground-truth shape $S_{GT}^{t}$ per frame using Procrustes analysis. We compute the average RMS 3D reconstruction error as $e_{3D} = \frac{1}{F}\sum_{t=1}^{F}\frac{\|S_{est}^{t} - S_{GT}^{t}\|_F}{\|S_{GT}^{t}\|_F}$. We used Kmeans++ algorithm \cite{arthur2007k} to initialize segments without disturbing the temporal continuity.

\begin{figure*}
\centering
\includegraphics[width=1.0\textwidth] {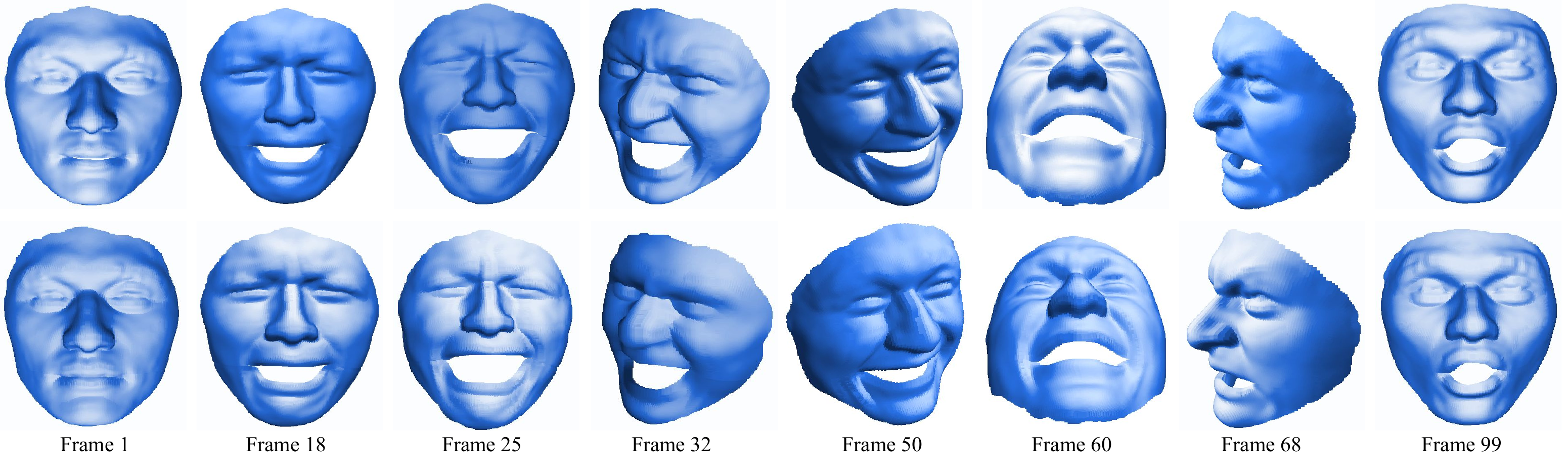}~~~
\caption{\small{Reconstruction results obtained on synthetic dense face dataset (face sequence 4). {\bf{Top row}} : Ground-truth 3D points, {\bf{Bottom row}} : Recovered 3D points using our approach.}}
\label{fig:syntheticSeq4res}
\end{figure*}

\begin{figure*}
\subfigure [\label{fig:d1}] {\includegraphics[width=0.33\textwidth, height= 0.14\textheight]{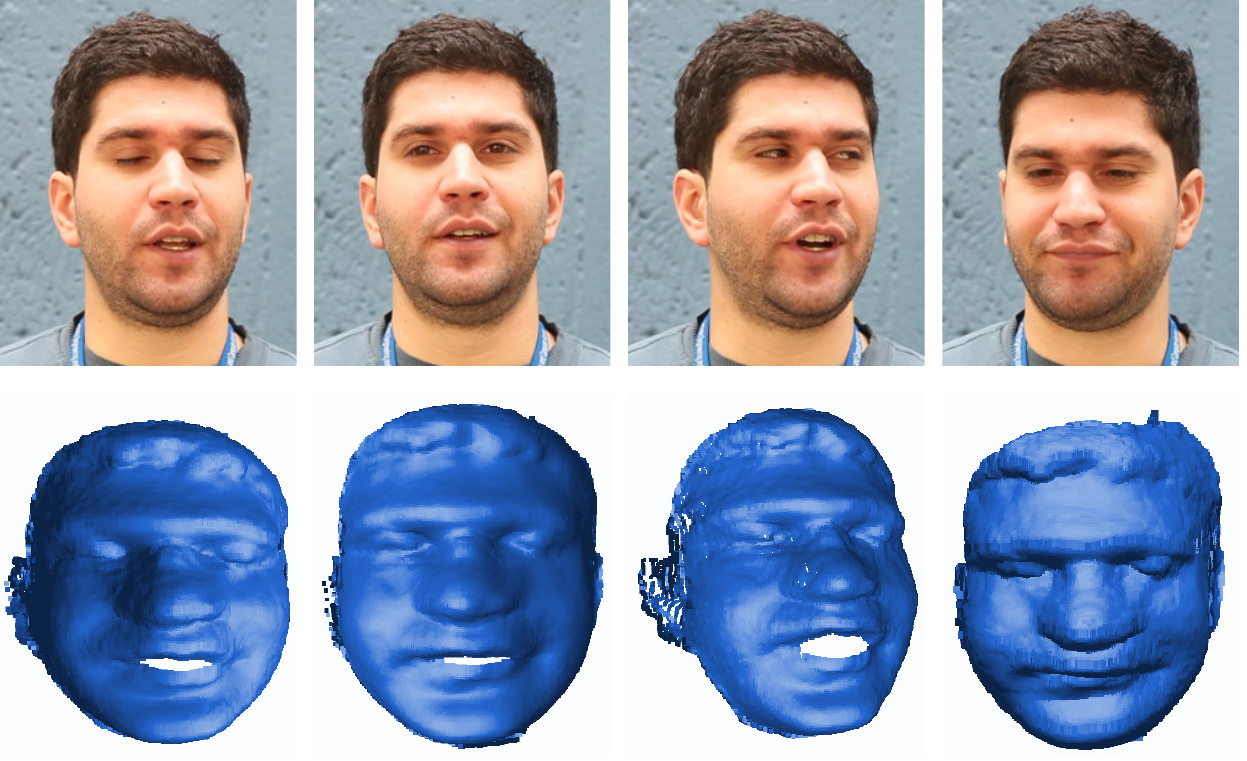}}
\subfigure [\label{fig:d2}] {\includegraphics[width=0.33\textwidth, height= 0.14\textheight]{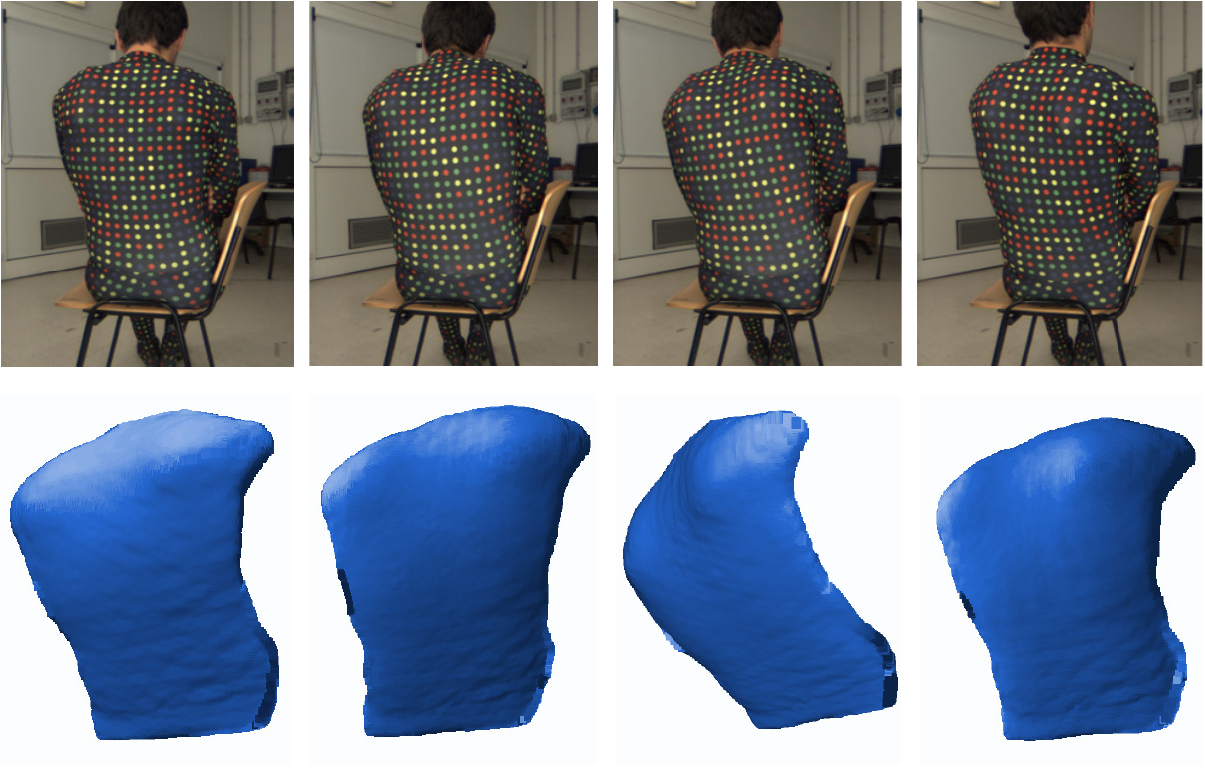}}
\subfigure [\label{fig:d3}] {\includegraphics[width=0.33\textwidth, height= 0.14\textheight]{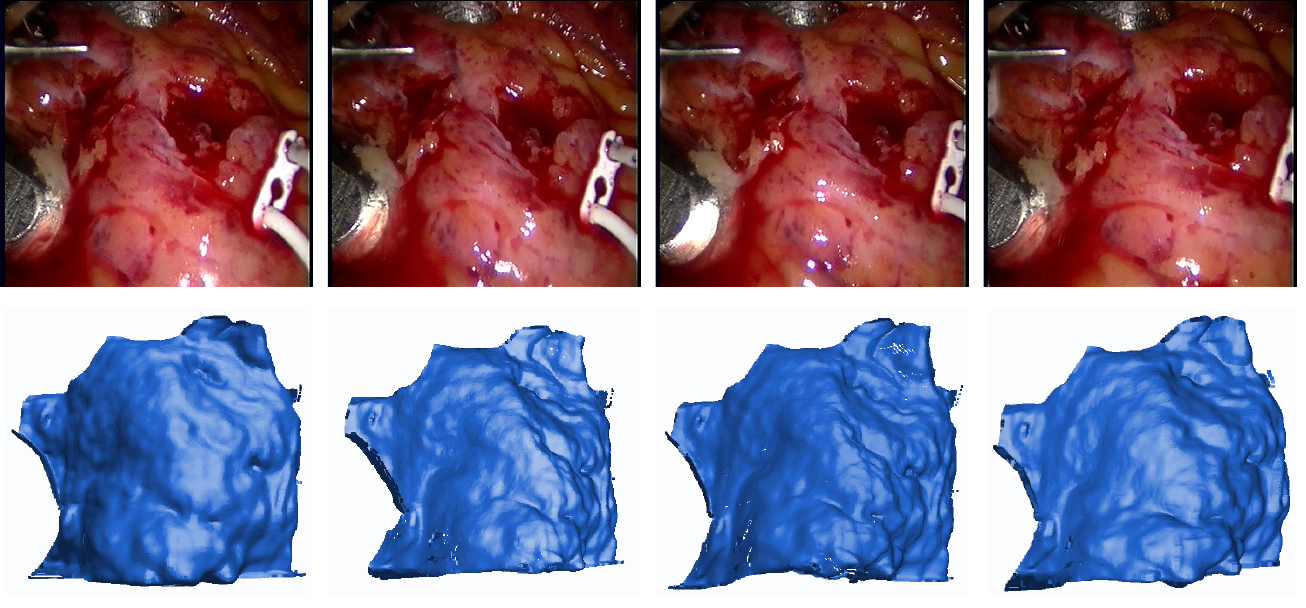}}
\caption{\small{Qualitative reconstruction results procured on benchmark real dense dataset \cite{garg2013dense} a) Face sequence (28,332 feature points over 120 frames)  b) Back sequence (20,561 feature points over 150 frames)  c) Heart sequence (68,295 feature points over 80 frames).}}
\label{fig:denseresults}
\end{figure*}

\begin{figure*}
\centering
\includegraphics[width=1.0\textwidth, height= 0.17\textheight] {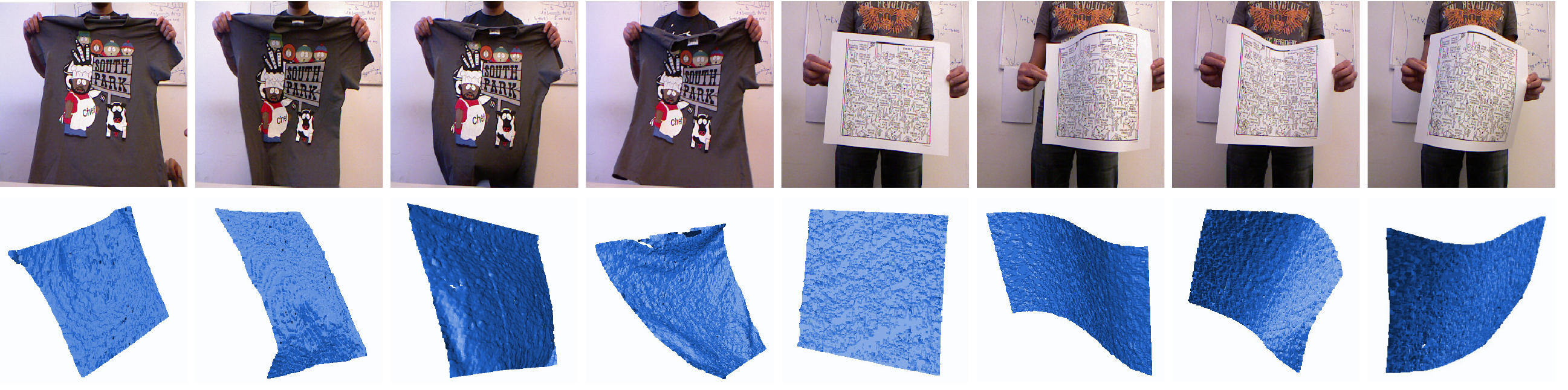}~~~
\caption{\small{Reconstruction results on benchmark kinect\_tshirt (74,000 points, 313 frames) and kinect\_paper(58,000 points, 193 frames) dataset \cite{varol2012constrained}. {\bf{Top row}}: Input image frame. {\bf{Bottom row}}: Dense 3D reconstruction for the corresponding frame using our approach.}}
\label{fig:kinect_paper_tshirt_res}
\end{figure*}

\noindent{\bf{Experiments on Synthetic Face Sequences:}} 
%This dataset is composed of 4 different face sequence of complex facial expression. These sequences contain 28,880 feature points tracked across 10 frames for face sequences \AC{unclear } 1, 2, and 99 frames for face sequence 3, 4. 
This dataset consists of 4 different face sequence with 28,880 feature points tracked over multiple frames. The face sequence 1, 2 is a 10 frame long video, whereas, face sequence 3, 4 is a 99 frame long video. It's a challenging dataset mainly due to different rotation frequencies and deformations in each of the sequence. Figure \ref{fig:syntheticSeq4res} shows the qualitative reconstruction results obtained using our approach in comparison to the ground-truth for face sequence 4. Table \ref{tab:densefacesequence} lists the performance comparisons of our method with other competing methods. Clearly, our algorithm outperforms the other baseline approach, which helps us to conclude that holistic approaches to rank minimization without drawing any inference from local subspace structure is a less effective framework to cope up with the local non-linearities.

%\AC{check.} 
\begin{table}[h]
\centering
\small
\begin{tabular}{c|c|c|c|c|l}
\hline
Data        & DS \cite{dai2017dense} & DV \cite{garg2013dense} & PTA \cite{akhter2009nonrigid} & MP \cite{paladini2012optimal} & Ours \\ \hline
Seq.1 &  0.0636   & 0.0531    &  0.1559   & 0.2572 & {\bf{0.0443}} \\ \hline
Seq.2 &  0.0569   & 0.0457    &  0.1503   & 0.0640 & {\bf{0.0381}} \\ \hline
Seq.3 &  0.0374   & 0.0346    &  0.1252   & 0.0611 & {\bf{0.0294}} \\ \hline
Seq.4 &  0.0428   & 0.0379    &  0.1348   & 0.0762 & {\bf{0.0309}} \\ \hline
\end{tabular}
\caption{ \small{Average 3D reconstruction error ($e_{3D}$) comparison on dense synthetic face sequence\cite{garg2013dense}. Note: The code for DV \cite{garg2013dense} is not publicly available, we tabulated its results from DS \cite{dai2017dense} work.}}\label{tab:densefacesequence}
\end{table}

\noindent{\bf{Experiments on face, back and heart sequence:}}  This dataset contains monocular videos of human facial expressions, back deformations, and beating heart under natural lighting conditions. The face sequence, back sequence, and heart sequence are composed of 28332, 20561, and 68295 feature points tracked over 120, 150, and 80 images, respectively. Unfortunately, due to the lack of ground-truth 3D data, we are unable to quantify the performance of these sequences. Figure \ref{fig:denseresults} shows some qualitative results obtained using our algorithm on this real dataset.

%The optimization variables can be obtained by solving the following sub-problems keeping all the constraints of Equation \eqref{eq:overallOptimization3} intact.

% \vskip -0.3cm
% \begin{equation}
% \begin{small}
% \begin{aligned}
% (S_s)^{k+1} = \underset{S_s}{\text{argmin}} ~\mathbf{E}((S_{s})^{k}, S_{t}^{\sharp}, C_s, C_t, J_s, J_t)
% \end{aligned}
% \end{small}
% \end{equation}

% \vskip -0.3cm
% \begin{equation}
% \begin{small}
% \begin{aligned}
% (S_t^{\sharp})^{k+1} = \underset{S_t^{\sharp}}{\text{argmin}} ~\mathbf{E}(S_{s}, (S_{t}^{\sharp})^{k}, C_s, C_t, J_s, J_t)
% \end{aligned}
% \end{small}
% \end{equation}

% \vskip -0.3cm
% \begin{equation}
% \begin{small}
% \begin{aligned}
% (C_s)^{k+1} = \underset{C_s}{\text{argmin}} ~\mathbf{E}(S_{s}, S_{t}^{\sharp}, (C_s)^{k}, C_t, J_s, J_t)
% \end{aligned}
% \end{small}
% \end{equation}

% \vskip -0.3cm
% \begin{equation}
% \begin{small}
% \begin{aligned}
% (C_t)^{k+1} = \underset{C_t}{\text{argmin}} ~\mathbf{E}(S_{s}, S_{t}^{\sharp}, C_s, (C_t)^k, J_s, J_t)
% \end{aligned}
% \end{small}
% \end{equation}

% \vskip -0.3cm
% \begin{equation}
% \begin{small}
% \begin{aligned}
% (J_s)^{k+1} = \underset{J_s}{\text{argmin}} ~\mathbf{E}(S_{s}, S_{t}^{\sharp}, C_s, C_t, (J_s)^{k}, J_t)
% \end{aligned}
% \end{small}
% \end{equation}

% \vskip -0.3cm
% \begin{equation}
% \begin{small}
% \begin{aligned}
% (J_t)^{k+1} = \underset{J_t}{\text{argmin}} ~\mathbf{E}(S_{s}, S_{t}^{\sharp}, C_s, C_t, J_s, (J_t)^{k})
% \end{aligned}
% \end{small}
% \end{equation}
%$k$ symbolizes the updates of the variables over iteration. 
\noindent{\bf{Experiments on kinect\_paper and kinect\_tshirt sequence:}} To evaluate our performance on the real deforming surfaces, we used kinect\_paper and kinect\_tshirt dataset\cite{varol2012constrained}. This dataset provides sparse SIFT\cite{lowe1999object} feature tracks along with dense 3D point clouds of the entire scene for each frame. Since, dense 2D tracks are not directly available with this dataset, we synthesized it. To obtain dense feature tracks, we considered the region within a window containing the deforming surface. Precisely, we considered the region within $xw$ = (253, 253, 508, 508), $yw$ = (132, 363, 363, 132) across 193 frames for paper sequence, and $xw$ =  (203, 203, 468, 468), $yw$ = (112, 403, 403, 112) across 313 frames for tshirt sequence to obtain the measurement matrix \cite{garg2013variational,garg2011robust}. Figure \ref{fig:kinect_paper_tshirt_res} show some qualitative results obtained using our method on this dataset. Table \ref{tab:kinectpapertshirtsequence} lists the numerical comparison of our approach with other competing dense NRSfM approaches on this dataset. 
% \footnote{Note: Few holes will still be present within the region of interest of dense point cloud, we did not consider those points and it was manually discarded for the experiment.} 
\begin{table}
\small
\centering
\begin{tabular}{c|c|c|c|c|l}
\hline
Data        & DS \cite{dai2017dense} & DV \cite{garg2013dense} & PTA \cite{akhter2009nonrigid} & MP \cite{paladini2012optimal} & Ours \\ \hline
paper & 0.0612   & -    &  0.0918   & 0.0827 & {\bf{0.0394}} \\ \hline
tshirt& 0.0636   & -    &  0.0712   & 0.0741 & {\bf{0.0362}} \\ \hline
\end{tabular}
\caption{ \small{Average 3D reconstruction error ($e_{3D}$) comparison on kinect\_paper and kinect\_tshirt \cite{varol2012constrained} sequence. Note: The code for DV \cite{garg2013dense} is not publicly available. The pixels with no 3D data available were discarded for the experiments and the evaluation.}}\label{tab:kinectpapertshirtsequence}
\vspace{-0.5cm}
\end{table}

\begin{figure*}
\centering
\subfigure [\label{fig:a1}] {\includegraphics[width=0.225\textwidth, height=0.13\textheight]{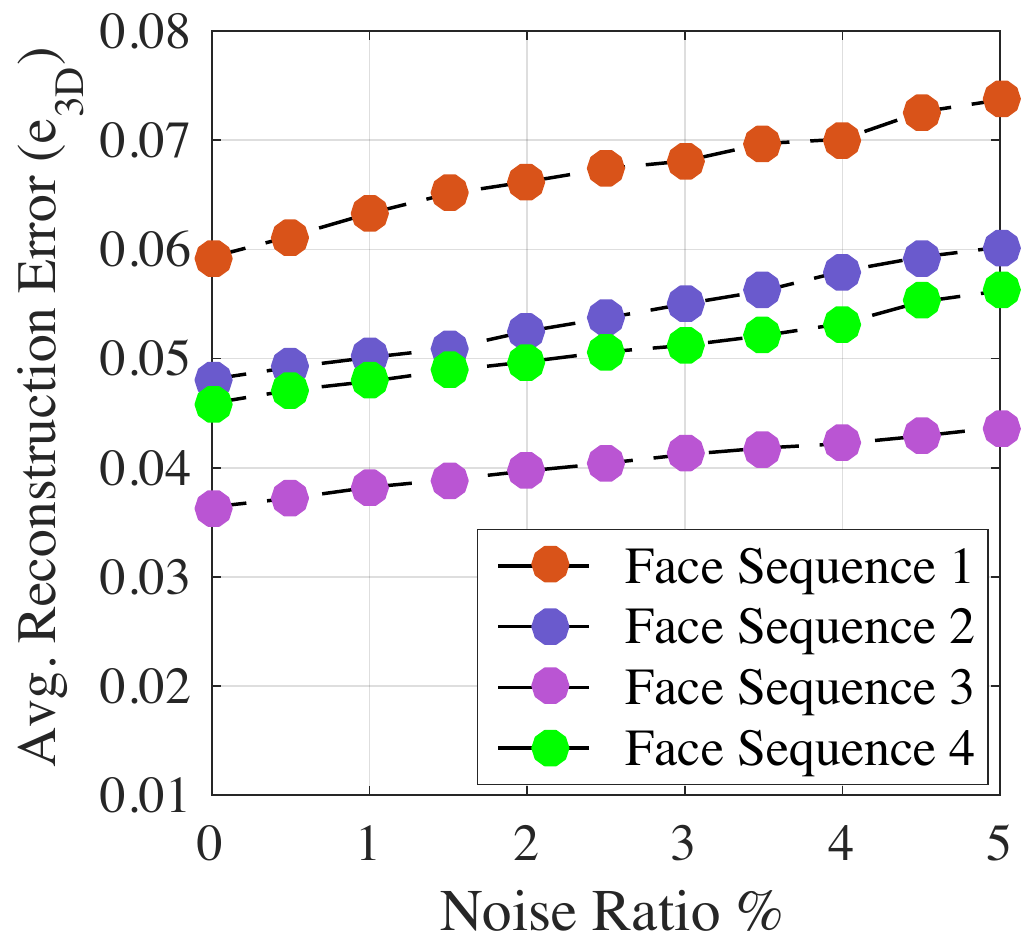}}
\subfigure [\label{fig:a2}] {\includegraphics[width=0.225\textwidth, height=0.13\textheight]{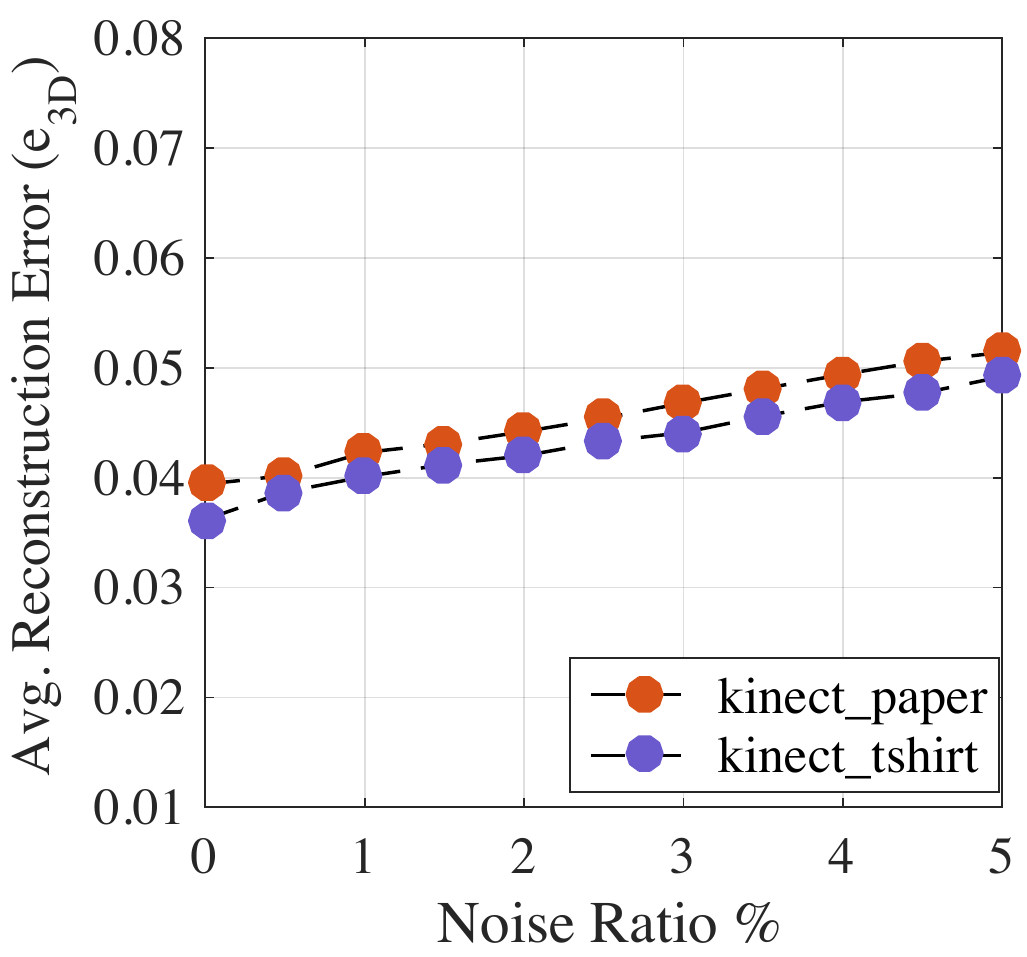}}
\subfigure [\label{fig:a3}] {\includegraphics[width=0.225\textwidth, height=0.13\textheight]{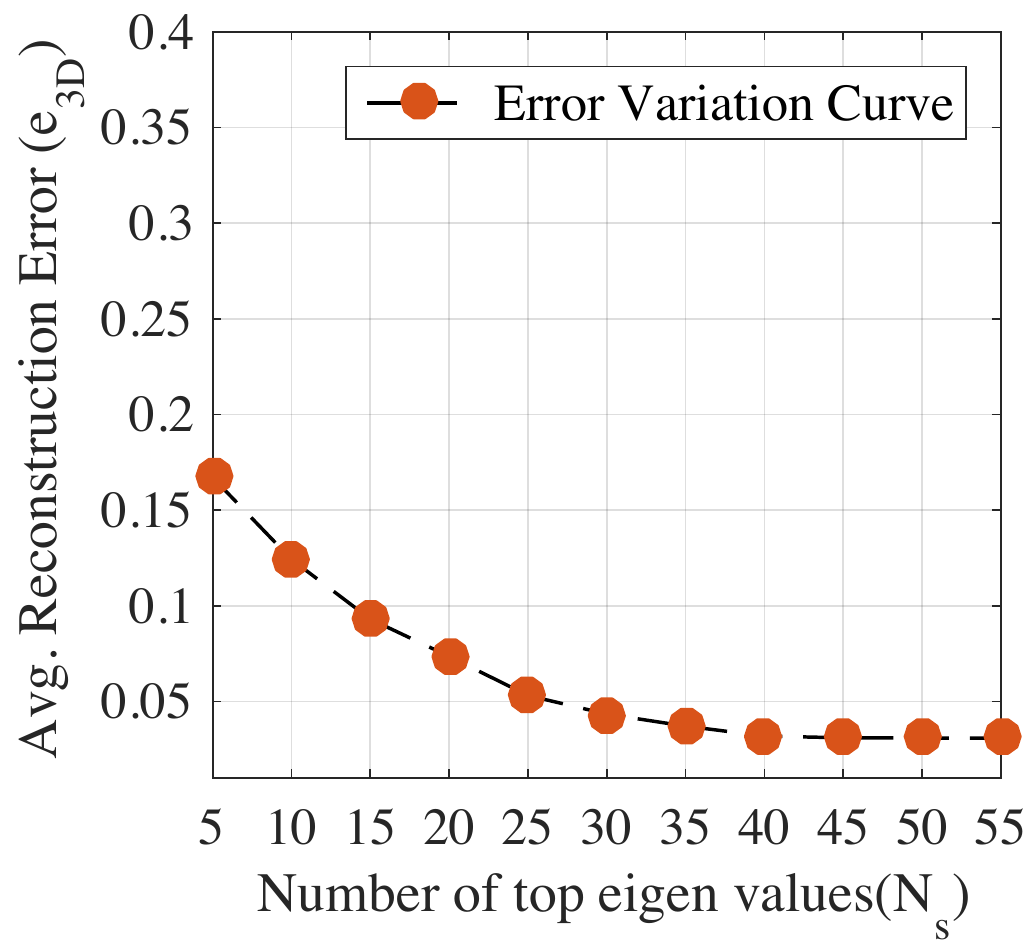}}
\subfigure [\label{fig:a4}] {\includegraphics[width=0.225\textwidth, height=0.13\textheight]{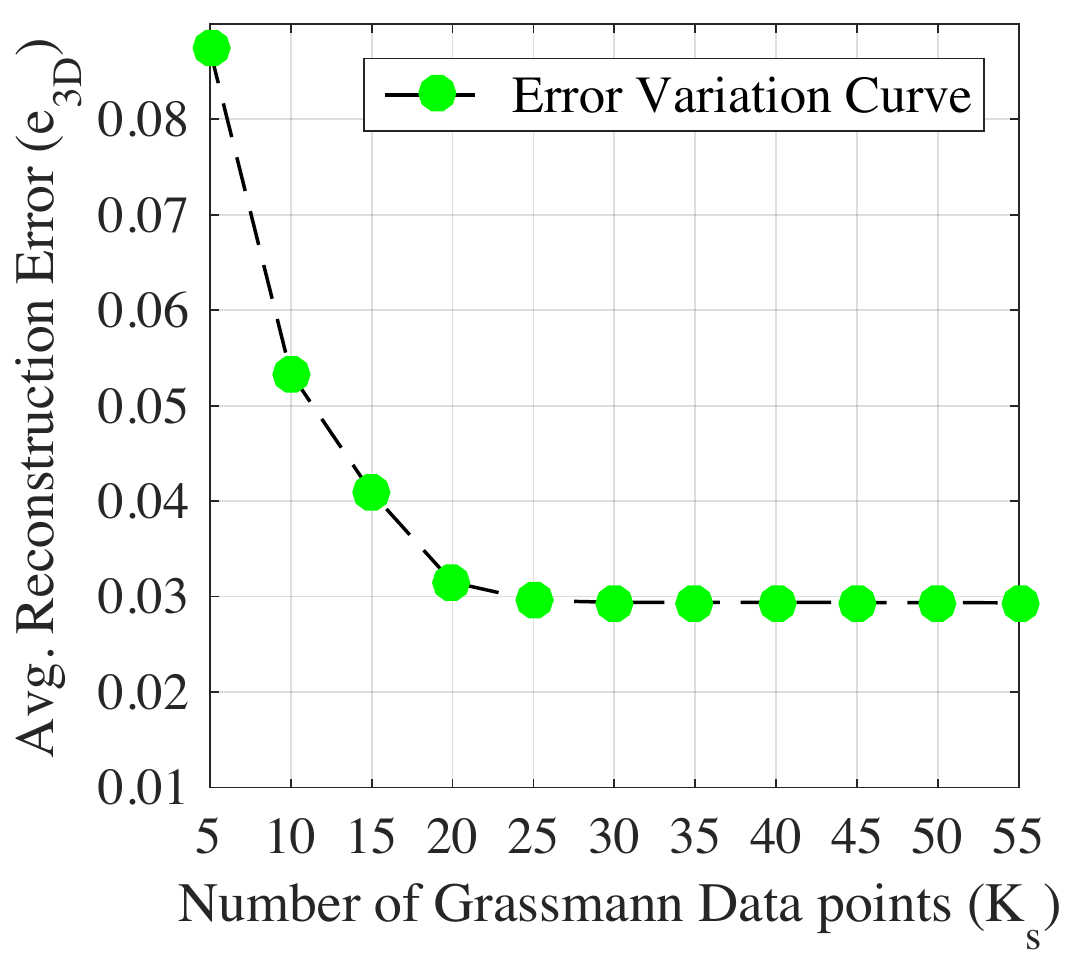}}
\caption{ \small{ (a)-(b) Avg. 3D reconstruction error ($e_{3D}$) variation with the change in the noise ratio for synthetic face sequence and kinect sequence respectively. (c)-(d)
Variation in $e_{3D}$ with the number of top eigen value and number of grassmann data points for Face Seq3.}}
\label{fig:noiseandNsVariation}
\end{figure*}

\begin{figure*}
\centering
\subfigure [\label{fig:l4}] 
{\includegraphics[width=0.22\textwidth, height=0.13\textheight]{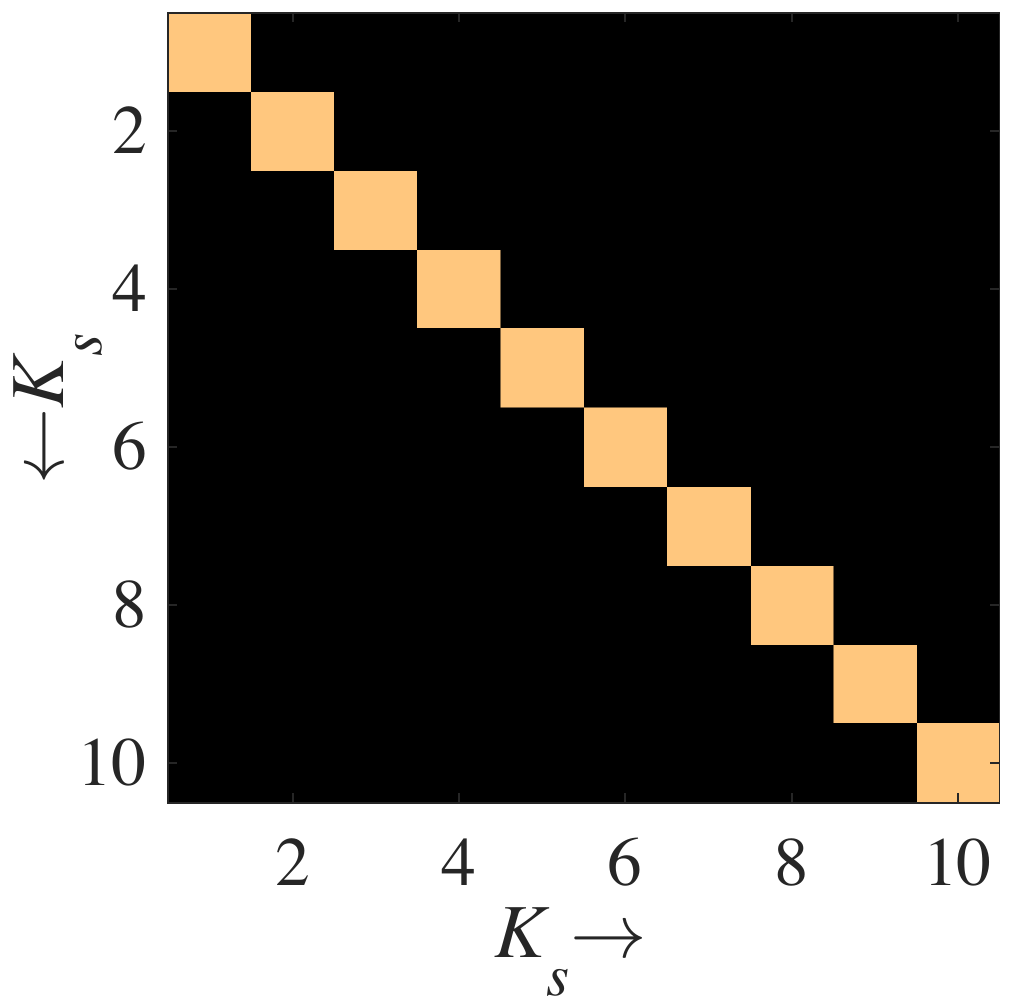}}
\subfigure [\label{fig:l3}] {\includegraphics[width=0.22\textwidth, height=0.13\textheight]{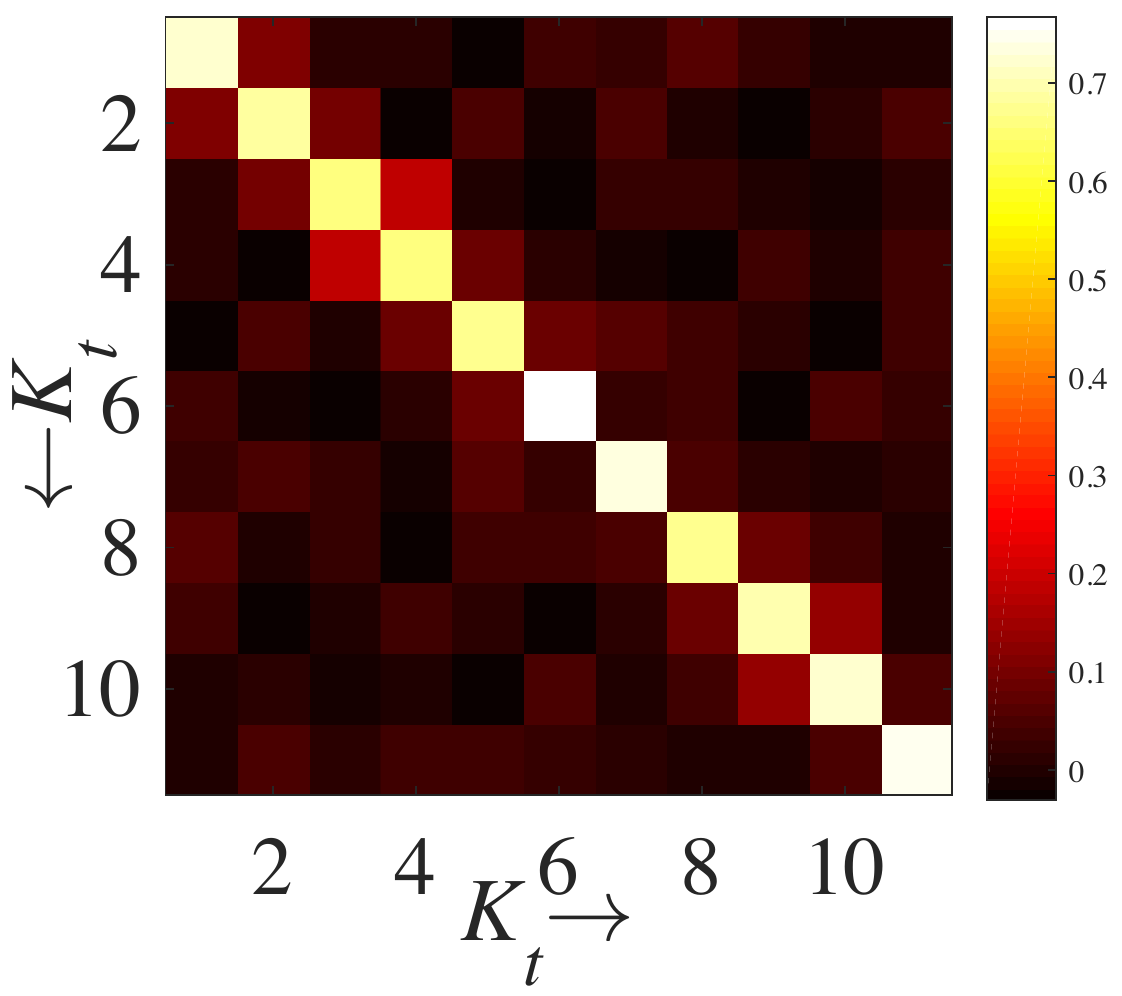}}
\subfigure [\label{fig:l2}] 
{\includegraphics[width=0.22\textwidth, height=0.13\textheight]{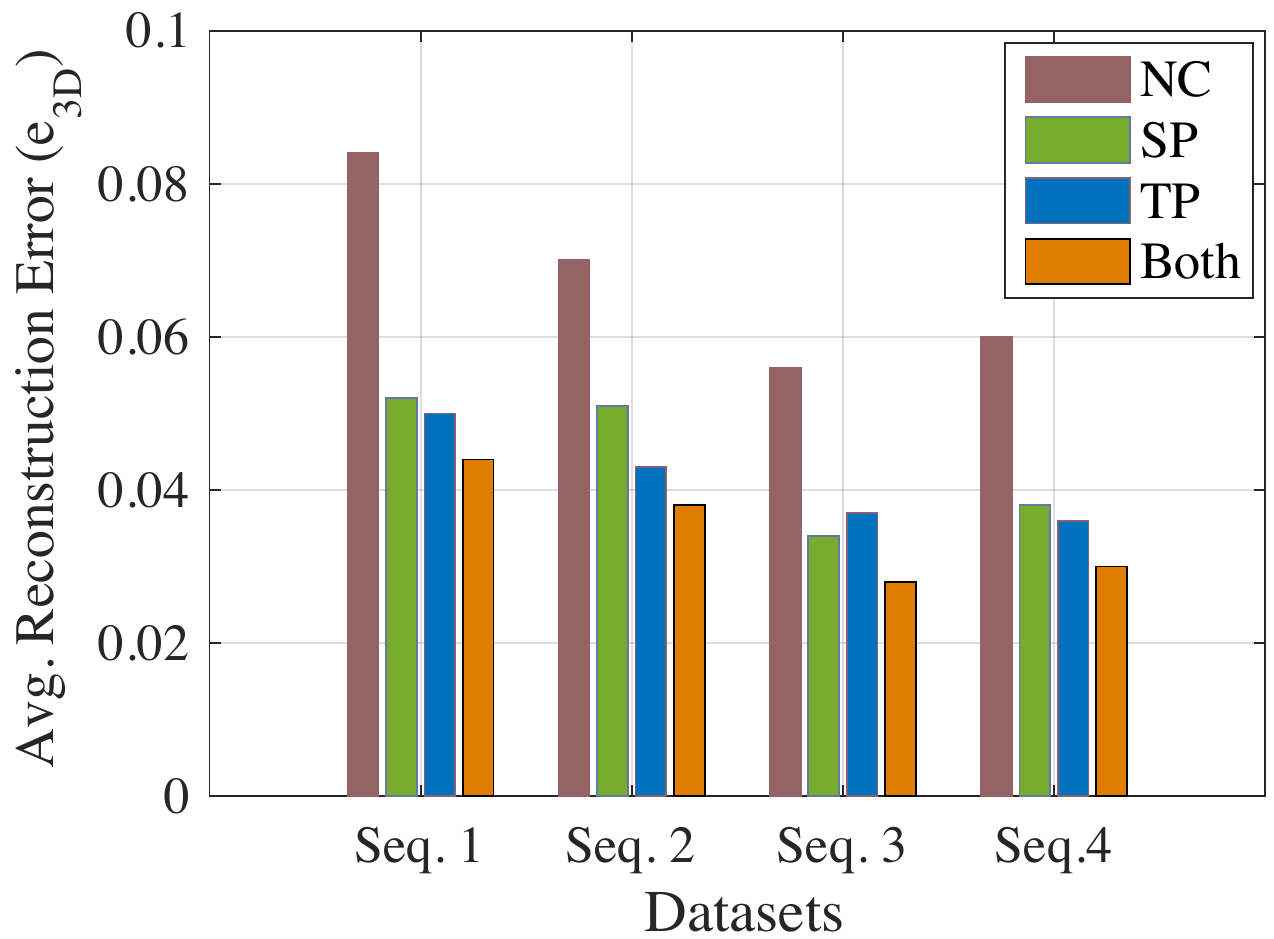}}
\subfigure [\label{fig:l1}] {\includegraphics[width=0.22\textwidth, height=0.13\textheight]{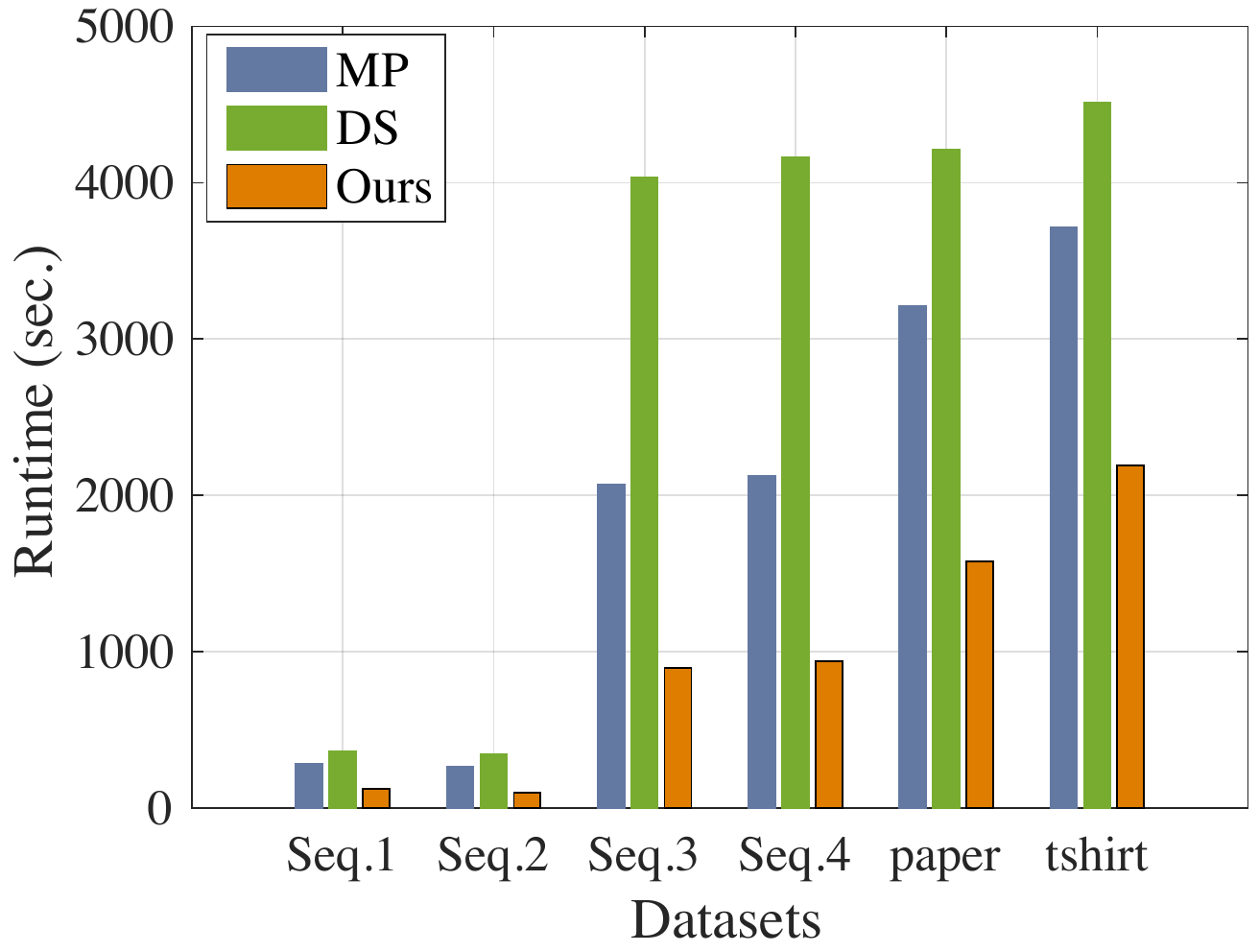}}
\caption{\small{(a)-(b) A typical structure of $C_s \in \mathbb{R}^{K_s\times K_s}$, $C_t \in \mathbb{R}^{K_t \times K_t}$ after convergence. (c) Ablation test performance on the synthetic face sequence \cite{garg2013dense}. (d) Runtime comparison of our method with MP \cite{paladini2012optimal} and a recent state-of-the-art dense NRSfM algorithm DS\cite{dai2017dense}.}}
\label{fig:timingsandCstructure}
\vspace{-0.3cm}
\end{figure*}
\noindent{\bf{Experiments on noisy data:}} To evaluate the robustness of our method to noise levels, we performed experiments by adding Gaussian noise under different standard deviations to the measurement matrix. Similar to DS \cite{dai2017dense} the standard deviations are incorporated as $\sigma_n = r\ \text{max}\{|W_s|\}$ by varying $r$ from 0.01 to 0.05. We repeated the experiment 10 times. Figure \ref{fig:a1} and Figure \ref{fig:a2} shows the variation in the performance of our method under different noise ratio's on synthetic face sequences\cite{garg2013dense} and kinect sequences\cite{varol2012constrained} respectively. It can be inferred from the plot that even with large noise ratios, the average reconstruction error does not fluctuate significantly. This improvement is expected from our framework as it is susceptible only to top eigen values.

\par{\bf{Effects of variable initialization on the overall performance:}} We  performed several other experiments to study the behavior of our algorithm under different variable initializations. For easy exposition, we conducted this experiment on noise free sequences. We mainly investigated the behavior of $N_s, N_t, K_s, K_t$ on the overall performance of our algorithm. Figure \ref{fig:a3} and Figure \ref{fig:a4} shows the variations in the reconstruction errors with respect to $N_s$ and $K_s$ respectively. A similar trend in the plots is observed for changes on $N_t$ and $K_t$ values. These plots clearly illustrate the usefulness of our local low-rank structure \ie, considering a small number of eigenvalues for every local structure is as good as considering all eigenvalues. Similarly, increasing the number of local subspaces after a certain value has negligible effect on the overall reconstruction error. Furthermore, we examined the form of $C_s$ and $C_t$ after convergence as shown Figure \ref{fig:l4} and Figure \ref{fig:l3}. Unfortunately, due to the lack of ground-truth data on local subspaces, we could not quantify $C_s$ and $C_t$. For qualitative analysis on the observation, kindly refer to the supplementary material. 

\par{\bf{Ablation Analysis:}} This test is performed to evaluate the importance of spatial and temporal constraints in our formulation. To do this, we observe the performance of our formulation under four different setups: a) without any spatio-temporal constraint (NC), b) with only spatial constraint (SP),  c) with only temporal constraint (TP), and d) with spatio-temporal constraint (Both). Figure \ref{fig:l2} shows the variations in reconstruction errors under these setups on four synthetic face sequence. The statistics clearly illustrate the importance of both constraints in our formulation. 

\par{\bf{Runtime Analysis:}} This experiment is performed on a computer with an Intel core i7 processor and 16GB RAM. The script to compute the runtime is written in MATLAB 2016b. Figure \ref{fig:l1} shows the runtime comparisons of our approach with other dense NRSfM methods. The runtime reported in Figure \ref{fig:l1} corresponds to the results listed in Table \ref{tab:densefacesequence}, \ref{tab:kinectpapertshirtsequence}. The results clearly show the \emph{scalability} of our method on datasets with more than 50,000 points. Despite PTA \cite{akhter2009nonrigid} is faster than our approach, its accuracy suffers by a large margin for dense NRSfM (see Table \ref{tab:densefacesequence}, \ref{tab:kinectpapertshirtsequence}).
% To analyze the runtime performance of our approach, we used synthetic face, real paper, and tshirt sequence.
% \textcolor{blue}{Suryansh: Compare the timings of your method with other methods at least on 4-5 datasets.}
% \subsection{Variation in 3D reconstruction error w.r.t noise}
% \textcolor{blue}{Face: Done. Kinect Paper and T-shirt: Done}
% \subsection{Variation in 3D reconstruction error w.r.t $N_s$ and $N_t$}
% \textcolor{blue}{Done}
% \subsection{Variation in 3D reconstruction error w.r.t number of grassmann data points($K_s$), $(K_t)$}
% \textcolor{red}{Pending}
% \subsection{Variation in timing w.r.t number of grassmann data points}
% \textcolor{red}{Pending}
% \subsection{Variation in reconstruction error w.r.t rank chosen for rotation matrix.}
% \textcolor{red}{Pending}
%\vspace*{-0.1em}

% \section{Conclusion}
% In this paper, we have introduced a scalable dense NRSfM algorithm which efficiently models the complex non-linear deformations. We achieved this by exploiting the non-linearity on the Grassmann manifold via spatio-temporal formulation. Moreover, we provided an efficient ADMM \cite{boyd2011distributed} based solution for solving our optimization. Several experiments on benchmark datasets are provided,  which clearly show the usefulness of our method. The proposed algorithm provides a new insight to model dense NRSfM which previously seems inconceivable. We believe that in practice, such a framework will be helpful to interesting 3D-vision applications. A natural extension of this work would be to address dense NRSfM for multi-body case \cite{kumar2016multi}.

\section{Conclusion}
In this paper, we have introduced a scalable dense NRSfM algorithm which efficiently models the complex non-linear deformations. We achieved this by exploiting the non-linearity on the Grassmann manifold via a spatio-temporal formulation. Moreover, we provided an efficient ADMM \cite{boyd2011distributed} based solution for solving our optimization. In the future, we will consider how to extend this work to the projective setting with perspective cameras (\eg \cite{projectivePAMI,projective}).  

\noindent\textbf{Acknowledgement.}{\footnotesize ~~ S. Kumar was supported in part by Australian Research Council (ARC) grant (DE140100180). Y. Dai was supported in part by National 1000 Young Talents Plan of China, Natural Science Foundation of China (61420106007), and ARC grant (DE140100180). H. Li is funded in part by ARC Centre of Excellence for Robotic Vision (CE140100016)}.

% \textcolor{blue}{Suryansh: My version of the paper is complete. Kindly, go through it (IF YOUR TIME PERMITS) and let me know your feedback. I thank you all for continuous support to get to this draft. Will the paper get through or not is a different story, but I am personally satisfied by giving a closure to what I started 2 years back. Thank you all.}
% [\textcolor{red}{ I feel THE FOLLOWING LAST TWO SENTENCES ARE MEANINGLESS AND USELESS.  LET US SAY SOEMTHING THAT ARE MORE CONCRETE AND MORE INSTRUCTIVE. FOR EAMXMPLE, WHAT OEN CAN KEARNB FROM THE PAPER, what are antural next step ?? rather than "Holy grail ...."  }]
% \textcolor{blue}{Suryansh: Hongdong, you are right, I have written a very weak conclusion and will rewrite it. Thanks for the comment. }
% We believe that, in practice such a framework will be helpful to many interesting real world applications that pivot around deformable shape reconstruction. We hope that such framework can be useful in solving the holy grail of NRSfM i.e multi-body dense NRSfM.
\balance
{\small
\bibliographystyle{ieee}
\bibliography{egbib}
}
%rough work
% Fluctuation in the reconstruction error w.r.t the variation in the noise level b) Performance variation with respect to change in the $N_s$ value. It clearly illustrate the point that with only few top eigen value and corresponding eigen vectors, we can achieve robust performance, thereby overcoming redundant computation and noisy signals. c)-d) The $C_s$ and $C_t$ matrix obtained for synthetic face sequence 3 after convergence. Kindly, note that $C_s$ and $C_t$ are the coefficient matrix of the subspaces not the vectorial points.

\end{document}